\documentclass[journal,twoside,web]{ieeecolor}
\usepackage{tmi}
\usepackage{graphicx}
\usepackage{subfigure}
\usepackage{amsmath}
\usepackage{multirow}
\usepackage{comment}
\usepackage{ulem}
\usepackage{url}

\def\etal{\normalem\emph{et al.~}}
\def\ie{\normalem\emph{i.e.}}
\def\eg{\normalem\emph{e.g.}}
\def\etc{\normalem\emph{etc.}}
\def\vs{\normalem\emph{vs. }}

\begin{document}
%
\title{Learning Hierarchical Attention for Weakly-supervised Chest X-Ray Abnormality Localization and Diagnosis}
%
%
%

\author{Xi Ouyang, Srikrishna Karanam,~\IEEEmembership{Member,~IEEE,} Ziyan Wu$^{*}$,~\IEEEmembership{Member,~IEEE,} Terrence Chen,~\IEEEmembership{Senior Member,~IEEE,} Jiayu Huo, Xiang Sean Zhou, Qian Wang$^{*}$, Jie-Zhi Cheng$^{*}$
\thanks{This work was supported in part by the National Key Research and Development Program of China (2018YFC0116400), Grant of Shanghai Strategic Emerging Industries from Shanghai Municipal Development and Reform Commission (20191211), and STCSM (19QC1400600). (Corresponding authors: Ziyan Wu, Qian Wang, and Jie-Zhi Cheng.)}
\thanks{Xi Ouyang, Jiayu Huo and Qian Wang are with the Institute for Medical Imaging Technology, School of Biomedical Engineering, Shanghai Jiao Tong University, Shanghai, China. Xi Ouyang and Jiayu Huo are interns at Shanghai United Imaging Intelligence Co. during this work. (e-mail: \{xi.ouyang,  jiayu.huo, wang.qian\}@sjtu.edu.cn).}
\thanks{Srikrishna Karanam, Ziyan Wu, and Terrence Chen are with United Imaging Intelligence, Cambridge MA, United States. (e-mail: \{srikrishna.karanam, ziyan.wu, terrence.chen\}@united-imaging.com).}
\thanks{Xiang Sean Zhou, and Jie-Zhi Cheng are with Shanghai United Imaging Intelligence Co., Ltd., Shanghai, China. (e-mail: sean.zhou@united-imaging.com, jzcheng@ntu.edu.tw).}
}

%
%

\markboth{IEEE TRANSACTIONS ON MEDICAL IMAGING}%
{Ouyang \MakeLowercase{\textit{et al.}}: Learning Hierarchical Attention for Weakly-supervised Chest X-Ray Abnormality Localization and Diagnosis}
%



\maketitle

\begin{abstract}
We consider the problem of abnormality localization for clinical applications. While deep learning has driven much recent progress in medical imaging, many clinical challenges are not fully addressed, limiting its broader usage. While recent methods report high diagnostic accuracies, physicians have concerns trusting these algorithm results for diagnostic decision-making purposes because of a general lack of algorithm decision reasoning and interpretability. One potential way to address this problem is to further train these models to localize abnormalities in addition to just classifying them. However, doing this accurately will require a large amount of disease localization annotations by clinical experts, a task that is prohibitively expensive to accomplish for most applications. In this work, we take a step towards addressing these issues by means of a new attention-driven weakly supervised algorithm comprising a hierarchical attention mining framework that unifies activation- and gradient-based visual attention in a holistic manner. Our key algorithmic innovations include the design of explicit ordinal attention constraints, enabling principled model training in a weakly-supervised fashion, while also facilitating the generation of visual-attention-driven model explanations by means of localization cues. On two large-scale chest X-ray datasets (NIH ChestX-ray14 and CheXpert), we demonstrate significant localization performance improvements over the current state of the art while also achieving competitive classification performance. Our code is available on \textcolor{cyan}{\url{https://github.com/oyxhust/HAM}}.
\end{abstract}


\IEEEpeerreviewmaketitle

\section{Introduction}

\IEEEPARstart{T}{he} chest X-ray (CXR) is one of the most commonly performed medical imaging examinations in clinical practice. Diagnosis with CXR greatly depends on the radiologist's experience \cite{kelly2016Radiology} since anatomical structures may overlap due to the 2D projection effect. Another challenge with CXR is the high diversity of possible abnormalities and diseases. Consequently, many detection \cite{Lakhani2017Radiology}, diagnosis \cite{guendel2018learning,guan2018diagnose,yan2018weakly}, and triage \cite{annarumma2019automated,ouyang2019weakly} methods have been proposed to support computer-aided CXR diagnosis. 

The public release of the large-scale NIH Chest X-ray14 \cite{wang2017chestx} and CheXpert \cite{irvin2019chexpert} datasets, both including more than 100,000 images, has further fostered research in this field, with  many prominent techniques \cite{guendel2018learning,guan2018diagnose,yan2018weakly,ma2019cross,baltruschat2019comparison} formulating CXR image diagnosis as a multi-label classification problem. Despite their high classification accuracies, physicians find it difficult to interpret these ``black-box” models. Furthermore, as these models are trained with image-level annotations, they are unable to capture the large intra-class diversity in shapes, appearances, and sizes of different abnormalities. 
On the other hand, one may expect to mitigate these issues given sufficient abnormality localization annotations for training; however obtaining them is prohibitively expensive, particularly in medical applications. 
Consequently, low accuracies in abnormality localization as well as limited model interpretability have become key bottlenecks in wide adoption of these algorithms in clinical practice.

Existing methods in the literature propose ways to address these issues. For instance, in Li \etal \cite{li2018thoracic}, a weakly supervised approach (using a small number of box annotations) was employed for abnormality localization. Despite good progress in a few cases, the performance in most abnormalities (\protect\eg, ``Nodule", ``Mass", ``Atelectasis", \protect\etc) still largely remains quite low. In addition to a general lack of decision reasoning (or model explanations), the performance improvements with this line of work have come at the cost of reduced image-level abnormality diagnosis performance \cite{li2018thoracic}. This is because of a severe data imbalance between the small number (of the order of a few hundreds) of box-level localization annotations versus a much larger number (of the order of several hundreds of thousands) of images with image-level class labels. 
Thus, in this work, our key considerations are - (a) \textsl{can we provide an efficient means for explaining model decisions?} and (b) \textsl{how do we improve localization performance while also ensuring little/no classification performance reduction?}


Recent progress in convolutional neural network (CNN) attention modeling and learning \cite{zhou2016learning,selvaraju2017grad} has led to increased adoption of visual attention for model interpretability and explainability. Such extensions \cite{li2018tell,wang2018sharpen}, while generating attention priors and demonstrating applicability for classification tasks, are not directly applicable in our context due to the hierarchical nature of localizing a particular abnormality in a larger anomalous region in the image. 
To address this problem, we design an attention-driven learning framework that addresses two key drawbacks of existing methods- (a) precisely localizing abnormalities, especially for subtle classes like ``Nodule", and (b) addressing the imbalance problem of box-level and image-level label annotations in a holistic manner. Our intuition is that CNN visual attention, by being a weakly-supervised source of localization cues, provides a strong prior for learning generalizable models even with small quantities of box-level annotations. Such a framework will not only help improve localization performance but also, by means of attention, provide a way to visually explain model decisions, both of which are important to clinical deployment of deep learning models.

\begin{figure}
\centering
 \includegraphics[width=8.9cm]{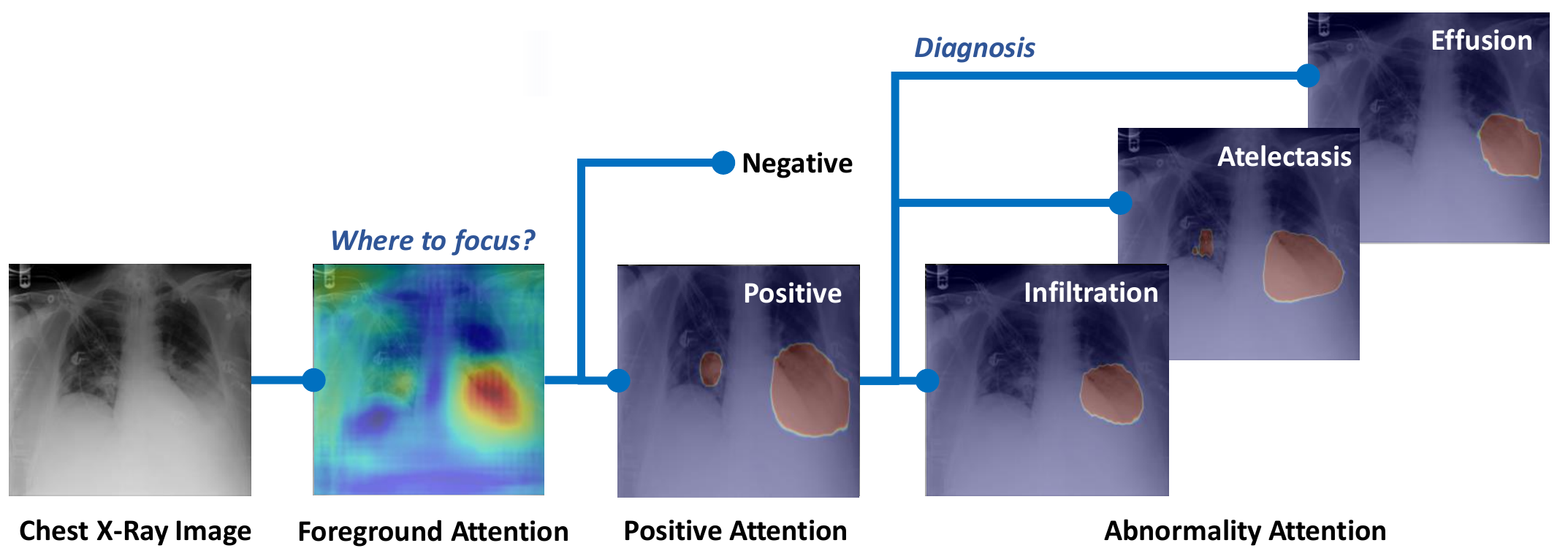}
\caption{Hierarchical attention mining framework. It contains three levels of attention mechanism: foreground attention, positive attention, and abnormality attention. The foreground attention is from the activation-based foreground attention block (FAB). The positive attention and abnormality attention are the gradient-based attentions generated from two online-CAM modules of the two-way (positive/negative) classification task and $D$-way ($D$ abnormality types) classification task, respectively.}
\label{fig:p1}
\end{figure}

Our proposed method learns hierarchical attention at three levels: foreground, positive, and abnormality attention (Fig. \ref{fig:p1}). To model foreground attention, we present an activation-based foreground attention block (FAB) that captures foreground dependencies to give an initial, coarse estimation of the foreground region of interest. 
Our FAB considers both channel- and position-wise attention, and is realized with a cascade design to guide the learning process to discover informative features that are useful for the later search/recognition of abnormalities.
Next, a two-way (positive/negative) classification scheme uses gradient-based diagnostic-level attention to specifically narrow down regions that may, again fairly coarsely, enclose the abnormalities (positive attention). Finally, abnormality attention from the $D$-way ($D$ abnormality types) classification task is responsible for computing the abnormality locations. To ensure learning of attention regions hierarchically, we enforce two explicit attention ordinality constraints. Specifically, we propose a novel attention-bound learning objective that enforces the output of abnormality attention to be located completely within the coarse positive attention region. Furthermore, we propose a novel attention-union objective to enforce positive attention to lie within the union region of abnormality attention maps. Finally, by design, our framework enables principled incorporation of the limited box-level annotations by regularizing the abnormality attention maps to directly conform to the ground-truth distribution. 

To summarize, the key contributions of our work are:
\begin{itemize}
    \item We present a new visual attention-driven weakly-supervised learning framework that simultaneously addresses abnormality localization and classification with very limited box-level annotations.
    \item To address the hierarchical nature of abnormality localization, our proposed visual attention mechanism is explicitly hierarchical and comprised of three levels (foreground, positive, and abnormality) that enables progressive weakly-supervised discovery of the specific abnormality location of interest.
    \item 
    We demonstrate improved abnormality localization performance, establishing state-of-the-art results on the NIH ChestX-ray14 dataset.
    \item We invite an experienced radiologist to provide box annotations in the CheXpert \cite{irvin2019chexpert} dataset to help evaluate our method's localization performance. In total, 2345 images in the CheXpert dataset have been annotated with 6099 bounding boxes for 9 abnormality types. 
\end{itemize}

\section{Related Work}


\subsection{CXR Image Analysis}
Deep learning has enabled much recent progress in the field of medical image analysis \cite{shen2017deep}. For CXR image analysis, the release of the NIH dataset \cite{wang2017chestx}, which provided an official patient-level data split, has motivated many recently studies \cite{guendel2018learning,guan2018diagnose,yan2018weakly,yao2018weakly,ma2019cross,baltruschat2019comparison} for the diagnosis of 14 abnormalities. Most of these techniques utilized either the DenseNet \cite{yao2018weakly,yan2018weakly,guan2018diagnose,guendel2018learning} or ResNet \cite{guan2018diagnose,baltruschat2019comparison} architecture as the backbone with several add-ons like squeeze-and-excitation \cite{yan2018weakly}, global-local feature branch combination \cite{guan2018diagnose}, two parallel branches \cite{ma2019cross}, integration of multi-resolution cues \cite{yao2018weakly}, knowledge fusion from other datasets \cite{guendel2018learning} and so on. The current state-of-the-art performance among methods that use the official data split \cite{guendel2018learning,yan2018weakly,yao2018weakly,ma2019cross,baltruschat2019comparison} is around 0.81 measured in terms of average AUC. Chen \etal \cite{chen2019deep} explored the hierarchical structure of the 14 abnormality labels from the PLCO dataset \cite{John2000The}.

\begin{figure*}
\centering
\includegraphics[width=0.98\textwidth]{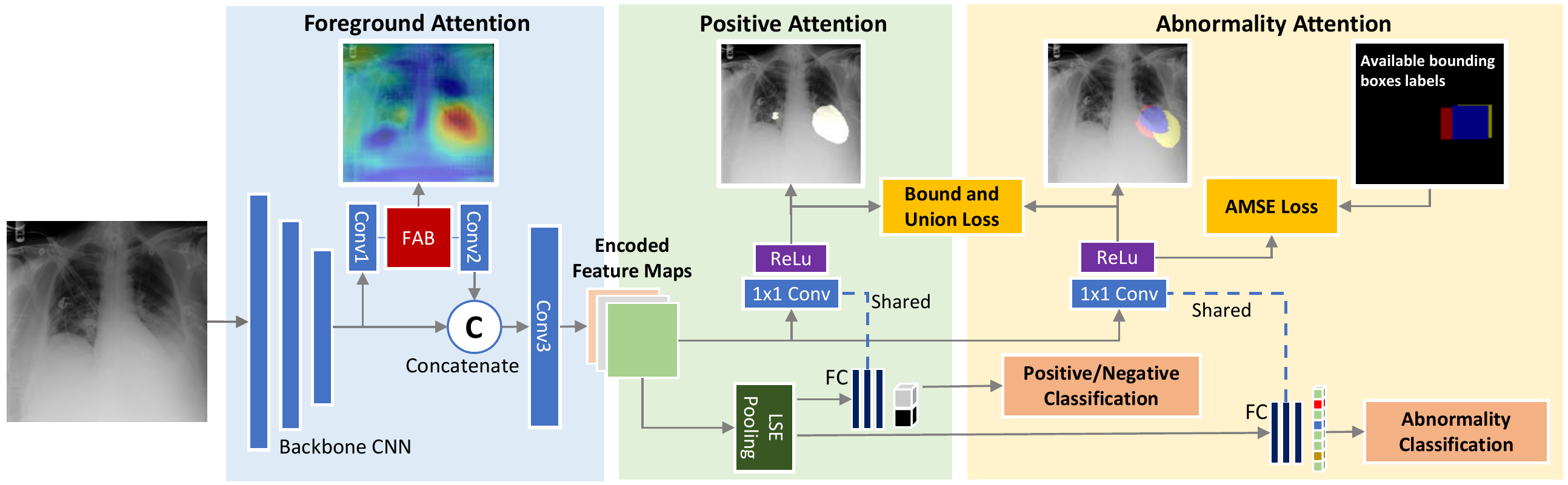}
\caption{Framework of our method. ``$1\times1$ Conv" denotes $1\times1$ convolutional kernel. ``FC" denotes the fully-connected layer. ``Conv1", ``Conv2", and ``Conv3" are the convolutional layers with $3\times3$ kernel. ``Conv1" reduces the number of channels of the feature maps from the backbone CNN for computation efficiency of FAB. ``Conv3" is used to concatenate the features of the backbone CNN and FAB, generating the encoded feature maps with 512 channels. The positive and abnormality attention results are generated with these encoded feature maps and the weights of two fully-connected layers from two-level classification tasks: the positive/negative classification and $D$-way abnormality classification task.}
\label{fig:framework}
\end{figure*}

As noted earlier, abnormality regions highlighted by models trained with image-level annotations may not always actually relate to the true abnormalities and may possibly also be non-pathological or outside the cardio-thoracic parts. To address these problems, some methods \cite{li2018thoracic,liu2019align} used a small number of available box-level annotations to facilitate more plausible abnormality localization. Specifically, multiple-instance learning (MIL) was employed in \cite{li2018thoracic,liu2019align,wang2019weakly} by treating the various spatial sliced blocks as instances.
While good localization was demonstrated in \cite{li2018thoracic}, the image diagnosis AUC was still around 0.75 with the official split (see Table 4 in the appendix of Li \emph{et al.}'s \cite{li2018thoracic} arXiv v6 version). Based on the MIL framework, Liu \etal \cite{liu2019align} developed the contrast-induced attention (CIA) network for localization improvement. CIA required pairs of positive and negative input images (images with and without abnormalities, respectively) to obtain contrast attention maps as the abnormality localization priors for the MIL framework. For the purpose of good quality contrast attention, CIA further required an alignment network to ensure that positive and negative images are in the same canonical form. While these techniques \cite{li2018thoracic,liu2019align} demonstrated better localization when compared to the NIH baseline \cite{wang2017chestx}, the performance on abnormalities like ``Nodule" and ``Mass" is still not satisfactory. 
Meanwhile, the performance of CXR diagnosis with the official split was not elaborated in \cite{liu2019align}. Our method, on the other hand, does not require either paired positive/negative images or the block slicing step as in these methods. Furthermore, our method improves abnormality localization performance without requiring an additional alignment step.

\subsection{Attention Mechanism}
Self-attention mechanism is an effective feature learning technique shown to be helpful in various image analysis tasks, \protect\eg, video classification \cite{wang2018non}, image classification \cite{hu2018squeeze}, and semantic segmentation \cite{fu2019dual}. This can be seen as a type of activation-based attention \cite{Zagoruyko2017AT} since a variety of activation functions, \eg, Sigmoid, Softmax, \etc, are used to compute attentive spatial parts or channels from feature maps. This is generally realized using two types of core modules: channel-wise attention \cite{hu2018squeeze} and spatial-wise attention \cite{wang2018non}, with both these types typically employed in a parallel fashion to address image analysis problems like semantic segmentation \cite{fu2019dual}. 

Gradient-based attention is another line of of work in this direction that was shown to be helpful for weakly supervised learning.
Class-activation map (CAM) \cite{zhou2016learning} and gradient-weighted class activation map (Grad-CAM) \cite{selvaraju2017grad,chattopadhay2018grad} are some examples of techniques that can be categorized under gradient-based attention, with CAM technically a special case of Grad-CAM for a specific type of CNN architecture (\ie, performing pooling over convolutional maps immediately prior to prediction). 

Some extensions \cite{fukui2019attention,li2018tell,wang2018sharpen} further used these technique as online trainable modules for improve the performance of models on downstream tasks, \eg, image classification or segmentation. There were some applications in the medical field as well, with Lian \etal \cite{mingxia1,mingxia2} utilizing gradient-based attention to boost the diagnostic performance of models for the Alzheimer’s disease (AD) in a two-stage framework. Specifically, the disease attention map derived in the first stage is employed to guide the training of the disease classification networks in the second stage. Since only AD-related diseases were considered in these methods \cite{mingxia1,mingxia2}, the aspect of label hierarchy, \eg, the image-level label of positive/negative \vs the abnormality-level labels was not fully explored. 

Our proposed method is, to the best of our knowledge, the first hierarchical visual attention framework that unifies both activation- and gradient-based attention in a holistic manner. To discover informative features that are useful for the later search/recognition of abnormalities, we propose the activation-based foreground attention block by cascading channel- and position-wise attention. Specifically, channel-wise attention module is used first to re-calibrate useful channels for the re-computation of feature maps. The recomputed feature maps may carry more informative features to support the learning of spatial attention features. With such a cascade design, effective attention features can be better learned. The resulting foreground attention map serves as a spatial prior for abnormal regions and thus helps the abnormality localization task. Furthermore, we develop two online CAM modules to produce gradient-based attention for positive and abnormality attention, given hierarchically organized image labels. Finally, we also propose novel ways to bound the learned attention by means of ordinal learning objectives to explicitly model the latent correlation between attention maps.

\section{Methodology}


Our framework is illustrated in Fig.~\ref{fig:framework}. It learns a three-level hierarchical representation of attention. The first level corresponds to a coarse delineation of the foreground region of interest, produced by our proposed foreground attention block (FAB). The FAB is realized by means of a differentiable activation-based attention mechanism that is infused into the model training process to highlight the features of the foreground region of interest. 
The second level corresponds to a coarse demarcation of the positive region of interest, realized by means of gradient-based attention with a two-way (positive/negative) classification scheme (called positive attention). Finally, the third level corresponds to delineating the particular abnormality of interest, realized using gradient-based attention with a $D$-way ($D$ abnormality types) classification scheme (called abnormality attention). To enforce explicit learning of such a hierarchical attention representation, we propose new ordinal attention constraints, implemented by means of learning objectives we call attention bound and attention union losses. Our learning mechanism is flexible to enable the use of a small number of box annotations, which helps improve localization performance without a reduction in the classification performance. In the subsequent sections, we discuss each component of our proposed framework in details.

\subsection{Foreground Attention Block (FAB)}

\begin{figure}
\centering
\includegraphics[width=8.5cm]{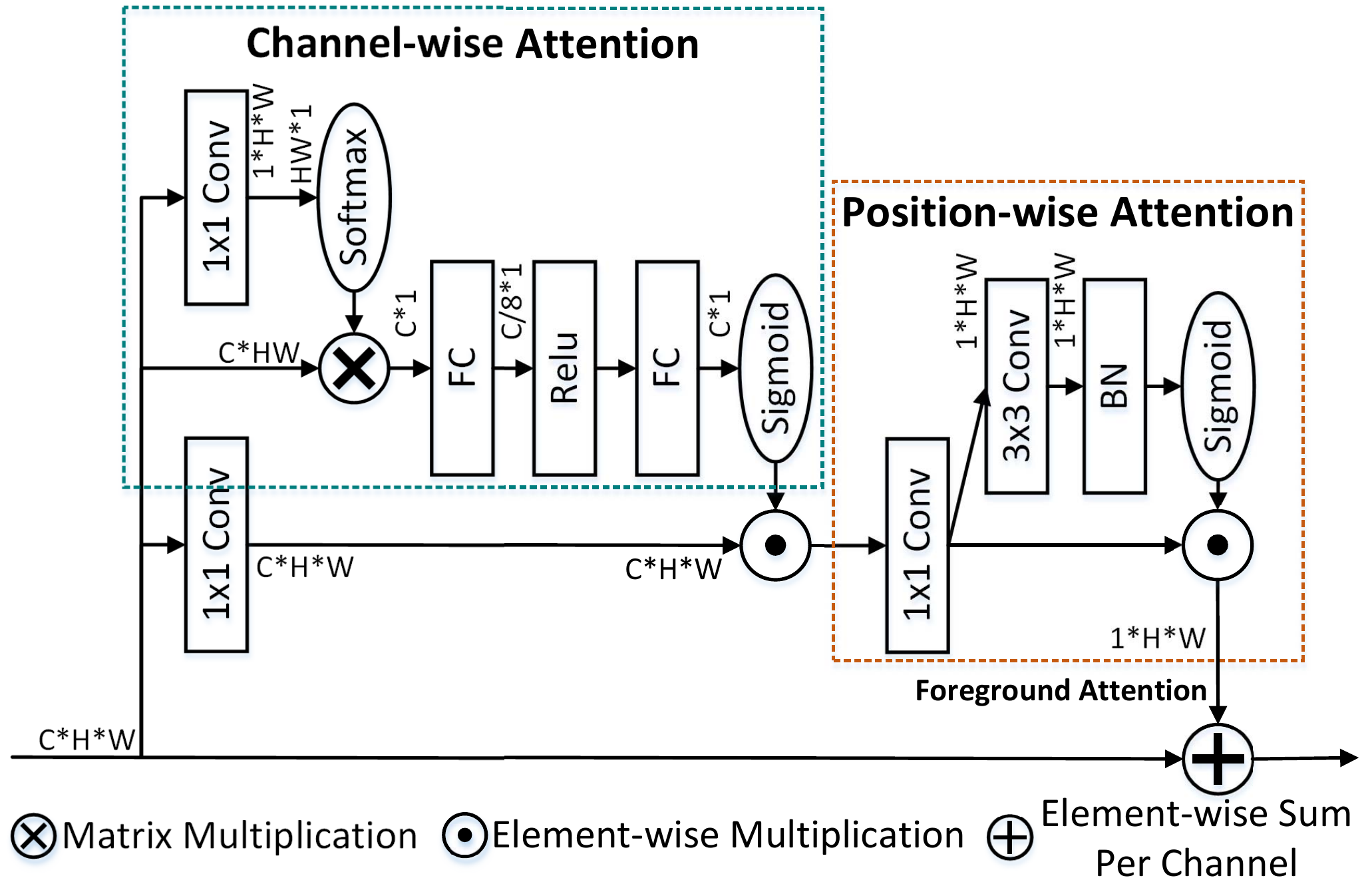}
\caption{Foreground Attention Block (FAB). ``BN" denotes batch normalization layer.
 We use the cascaded structure of channel- and position-wise attention to produce the foreground attention (shown in Fig. \ref{fig:framework}), which is element-wise added to each channel of the input $C\times H \times W$ feature maps.
}
\label{fig:attention}
\end{figure}

The foreground attention block (see Fig.~\ref{fig:attention}) implements perceptual-level attention with a self-attention mechanism \cite{wang2018non,hu2018squeeze,cao2019gcnet} to learn and highlight foreground features. The FAB is comprised of both channel and position attention modules, which we propose to use in a sequential manner. Unlike Fu \etal \cite{fu2019dual} that combined these two attention types into a parallel data flow, 
we cascade the channel- and position-wise attention to learn a foreground attention map. Specifically, the channel-wise attention is performed first to calibrate useful channels for the position-wise attention component.
Fu \etal \cite{fu2019dual} on the other hand exercised the self-attention mechanism in a parallel manner to recalibrate the channel and spatial cues simultaneously. Accordingly, the parallel self-attention mechanism with two non-local-based matrix operations may consume more GPU memory. Our cascading FAB only requires one matrix operation and can attain the desirable results in an efficient and faster way.

Our channel-wise attention module is architecturally similar to the squeeze-and-excitation (SE) block \cite{hu2018squeeze}, where we replace the average pooling operation with spatial attention pooling. Specifically, as shown in Fig. \ref{fig:attention}, we employ spatial attention \cite{wang2018non,cao2019gcnet} to assign similar weights to pixels having similar scores using softmax, which are then used to perform weighted spatial average pooling of the input feature maps, producing a channel-weighted $C \times $1 vector. The channel-weighted (by the $C \times $1 vector above) $C\times H \times W$ feature maps are then processed by the position-wise attention module, producing an attention-weighted $1 \times H \times W$ map. It is the foreground attention to give an initial, coarse estimation of the foreground region of interest. Then, it is element-wise added to each channel of the input $C\times H \times W$ feature maps, producing the final features that are then input to the subsequent parts of our model.

\subsection{Diagnostic Attention}
\textcolor{black}{After identifying the foreground regions of interest, our model computes diagnostic attention at the remaining two levels of the hierarchy, which is realized with different but coupled classification objectives. First, we perform a two-way classification and generate the positive attention map. This step essentially attempts to tell apart CXR images with and without any abnormalities, thereby helping learn features that are important for positive/negative prediction. Next, the same feature maps are used in conjunction with a $D$-way classifier, where the goal is to identify the particular abnormality type among the $D$ possibilities, to generate the abnormality attention maps. To exhaustively learn all features important for these classification tasks, and to produce the corresponding attention maps, we use a gradient-based attention mechanism, \protect\eg, online CAM \cite{zhou2016learning,fukui2019attention}. The key idea of CAM is to generate the attention maps for different classes by weighting the convolutional feature maps with the weights from the fully-connected layer. Let $f$ denote the feature maps before the \emph{log-sum-exp} (LSE) pooling \cite{sun2016pronet} operation and $w$ denote the weight matrix of the fully-connected layer. To make our attention generation procedure trainable, we use $w$ as the kernel of a $1\times1$ convolution layer such that:
\begin{equation}
\label{eq:cam}
M = {\rm{ReLU}}\left( {{\rm{conv}}\left( {f,w} \right)} \right),
\end{equation}
where $M$ has the shape $D \times T \times S$, and $D$ is the number of classes in the $D$-way ($D$ abnormality types) classification task. $D$ is set to $1$ for the the two-way (positive/negative) classification task. Given the attention map $M$, we normalize the values to the range between 0 and 1 and perform sigmoid for soft masking \cite{li2018tell}, which is defined as:
\begin{equation}
\label{eq:thresholding}
T(M) = \frac{1}{{1 + \exp ( - \alpha (M - \beta ))}},
\end{equation}
where values of $\alpha$ and $\beta$ are set to $100$ and $0.4$ respectively.}

\textcolor{black}{Based on Equations \ref{eq:cam} and \ref{eq:thresholding}, we use the encoded feature maps from the backbone CNN and the weights of fully-connected layer of the 2-way (positive/negative) classification task to generate the positive attention map $M^P$. The abnormality attention maps $M^{a_k}$ ($k=1,2,\ldots,D$ indicates the specific abnormality) are similarly computed with the encoded feature maps, which are further weighted with the weightings of fully-connected layer in the $D$-way classification task, see Fig. \ref{fig:framework}.
The synergy between the two classification tasks is achieved by means of our new ordinal constraints on the attention maps. Our intuition is that each abnormality attention map (obtained from the $D$-way classifier) should be completely contained within the positive attention map obtained from the 2-way (positive/negative) classification task. Furthermore, the positive attention map itself should be constrained to cover all the possible regions that may be attended to by the individual abnormality attention maps. To this end, we next formulate our new attention bound and attention union objective functions, which will be detailed later. Finally, we can exploit the small number of box-level annotations to further improve the localization performance in the weakly supervised setting. }

\subsubsection{The attention bound loss}
\begin{figure}[!t]
\centering
\subfigure[]{
\begin{minipage}{3.9cm}
\centering
\label{fig:relation:a}
\includegraphics[width=3.8cm]{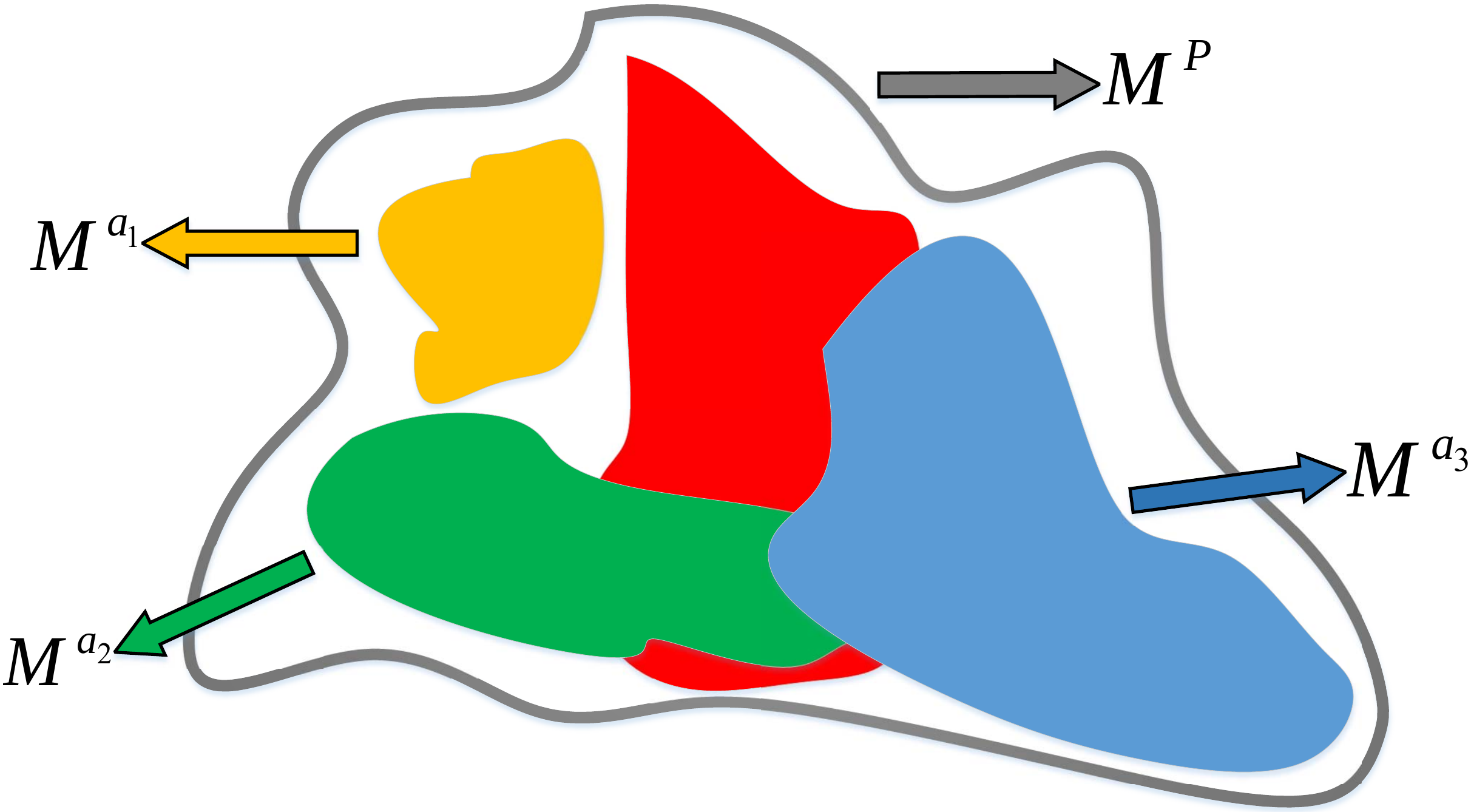}
\end{minipage}
}
\subfigure[]{
\begin{minipage}{3.9cm}
\centering
\label{fig:relation:b}
\includegraphics[width=3.8cm]{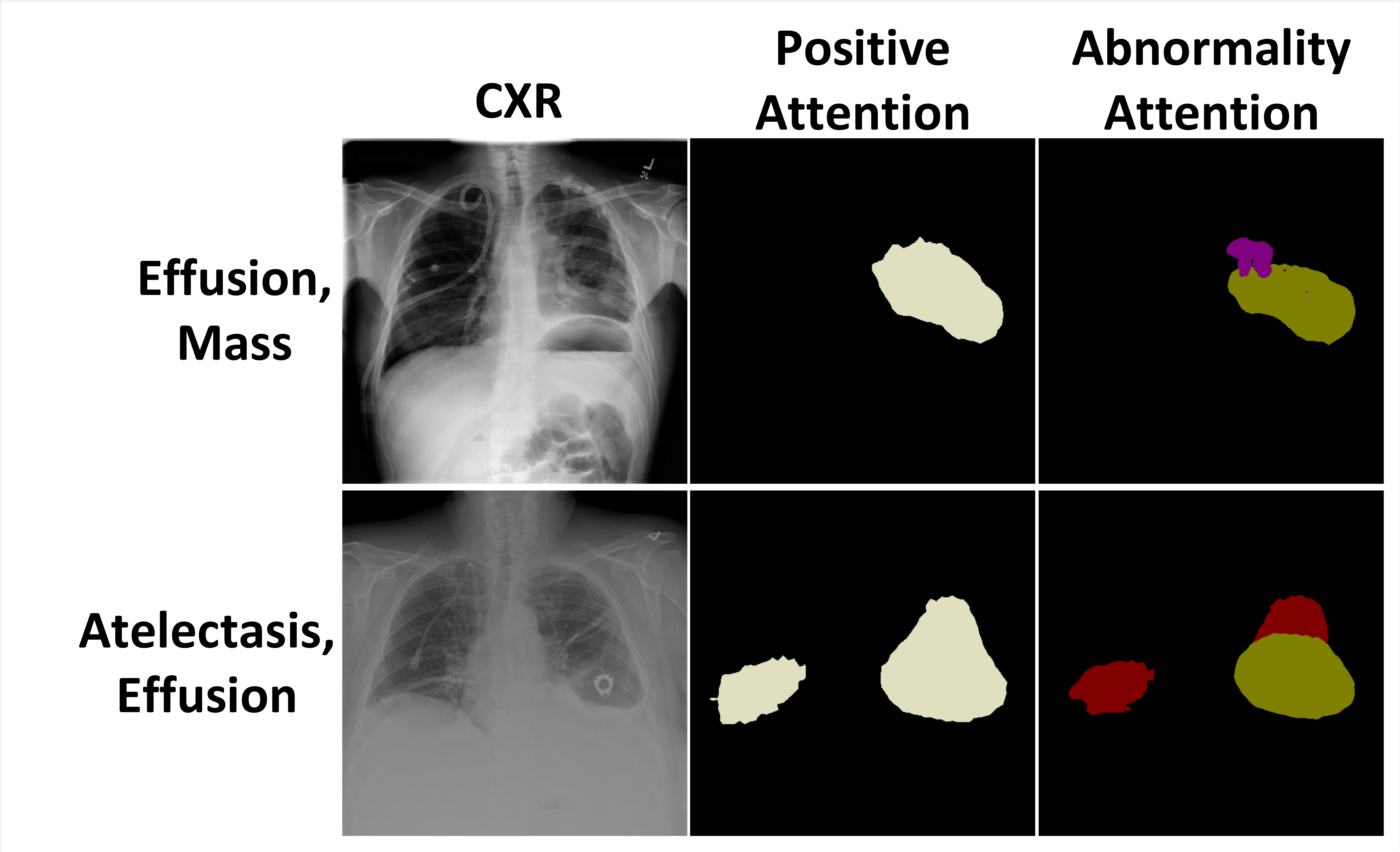}
\end{minipage}
}
 \caption{(a) Relation of positive attention map $M^{p}$ and abnormality attention maps $M^{a}$.
 (b) Real case illustration. Different abnormality attention maps $M^{a_k}$ are represented in different colors.} 
 \label{fig:relation}
\end{figure}

\textcolor{black}{As noted above, we generate two diagnostic attention maps in attention hierarchy: the positive attention map $M^{p}$ with shape $1 \times H \times W$ and $D$-way abnormality attention maps $M^{a_k}$ with shape $D \times H \times W$ (where $D$ is the number of abnormality classes and $k=1,2,\ldots,D$ indicates the specific abnormality). Given the hierarchical relationship between these two attention maps discussed above, we seek a learning objective that enables the model to learn to produce attention that respects this hierarchy. Specifically, given a CXR image, all abnormality attention maps $M^{a_k}$ shall be contained within the region bounded by the positive attention map $M^{p}$ as shown in Fig. \ref{fig:relation}. Accordingly, our proposed attention bound objective, $L_{bound}$, attempts to spatially constrain the region covered by each $M^{a_k}$ with respect to $M^{p}$, and is realized for input CXR image as:
\begin{multline}
\label{eq:bound}
{L_{bound}} = \\  \frac{1}{N}\sum\nolimits_{{y_k} = 1} {\left( {1 - \frac{{\sum\nolimits_{ij} {(\min (M_{ij}^p,M_{ij}^{{a_k}}) \cdot T(M_{ij}^{{a_k}}))} }}{{\sum\nolimits_{ij} {M_{ij}^{{a_k}}} }}} \right)},
\end{multline}
where the image's ground-truth label $y_k \in \{0,1\}$ ($0/1$ indicates absence/presence of abnormality $k$), $N$ is the number of positive classes in the image label set $\{y_k\}$, $i$ and $j$ represent the $(i,j)^{th}$ pixel in the corresponding attention map, and $T(M_{ij}^{{a_k}})$ is the soft masking operation defined in Equation \ref{eq:thresholding} which masks out the impact of background noise in the attention maps. In summary, the $L_{bound}$ can ensure abnormality attentions $M^{a}$ to lie within the positive attention $M_p$. }

\subsubsection{The attention union loss}
\textcolor{black}{While our proposed $L_{bound}$ helps enforce explicit spatial constraints on the spatial bounds of the abnormality attention $M^{a_k}$, it does not, however, provide any direct supervision on the spatial extent of the positive attention $M^p$. To ensure the positive attention only encompasses all the possible abnormality attention regions, and no more than that, we enforce explicit union constraint on the spatial extent of positive attention map $M^p$ and the union of all the abnormality attention maps $M^{a_k}$. Specifically, as shown in Fig. \ref{fig:relation}, we seek the spatial extent of the positive attention map $M^p$ to be no more than the spatial extent of the} union of all the abnormality attention maps $M^{a_k}$. To this end, we first compute the union of all abnormality attention maps, denoted $M^u$, as:
\begin{equation}
\label{eq:attention_union}
{M^u} = \max (M_{ij}^{{a_1}} \cdot {y_1},M_{ij}^{{a_2}} \cdot {y_2},\dots,M_{ij}^{{a_D}} \cdot {y_D}),
\end{equation}
where similar to $L_{bound}$, we only consider the positive abnormalities. Then, our proposed attention union loss, $L_{union}$, is realized by calculating the region overlap between $M^p$ and $M^u$ as:
\begin{equation}
\label{eq:union}
{L_{union}} = 1 - \frac{{\sum\nolimits_{ij} {(\min (M_{ij}^p,M_{ij}^{u}) \cdot T(M_{ij}^p) )} }}{\sum\nolimits_{ij} {M_{ij}^p}},
\end{equation}
where $T(M_{ij}^p)$ refers to the same soft-masking operation of Equation~\ref{eq:thresholding}.

\subsection{Weakly-supervised Learning from Extra Box-level Annotations}
Although we have proposed the attention bound and union loss to better locate the abnormality regions, it is still a challenging task under only image-level labels. Given a small number of extra annotations (\protect\eg, bounding-box annotations for abnormal localizations), we can in fact provide weak yet direct supervision to our model. To this end, we apply an attention-adaptive mean square error (AMSE) loss, $L_{amse}$.
Given our abnormality attention map $T({M}^{a_k})$ (bilinearly interpolated to input size after soft masking) and the corresponding ground-truth abnormality annotation $G^k$, our $L_{amse}$ is formulated as:
\begin{equation}
\label{eq:amse}
{L_{amse}} = \frac{1}{N}\sum\nolimits_{{y_k} = 1} {\left( {\frac{{\sum\nolimits_{ij} {{{(T(M_{ij}^{{a_k}}) - G_{ij}^k)}^2}} }}{{\sum\nolimits_{ij} {T(M_{ij}^{{a_k}}) + \sum\nolimits_{ij} {G_{ij}^k} } }}} \right)},
\end{equation}
where $N$ is the number of positive abnormalities in the image label set $\{y_k\}$. One can note that the proposed $L_{amse}$ is a slightly modified version of the traditional MSE loss using the sum of regions of location map ${M}^{a_k}$ and $G^k$ as an adaptive normalization factor. We show later that our $L_{amse}$ improves localization performance even with a small number of box annotations. In the datasets used in this work, only box-level annotations are available. To get the ground-truth masks, for the $k^{th}$ abnormality class, given the box annotation, we first generate a pixel-level binary mask $G^k$ where all pixels outside the current bounding box location are set to 0.

\begin{figure}
\centering
\includegraphics[height=3.6cm, width=8.2cm]{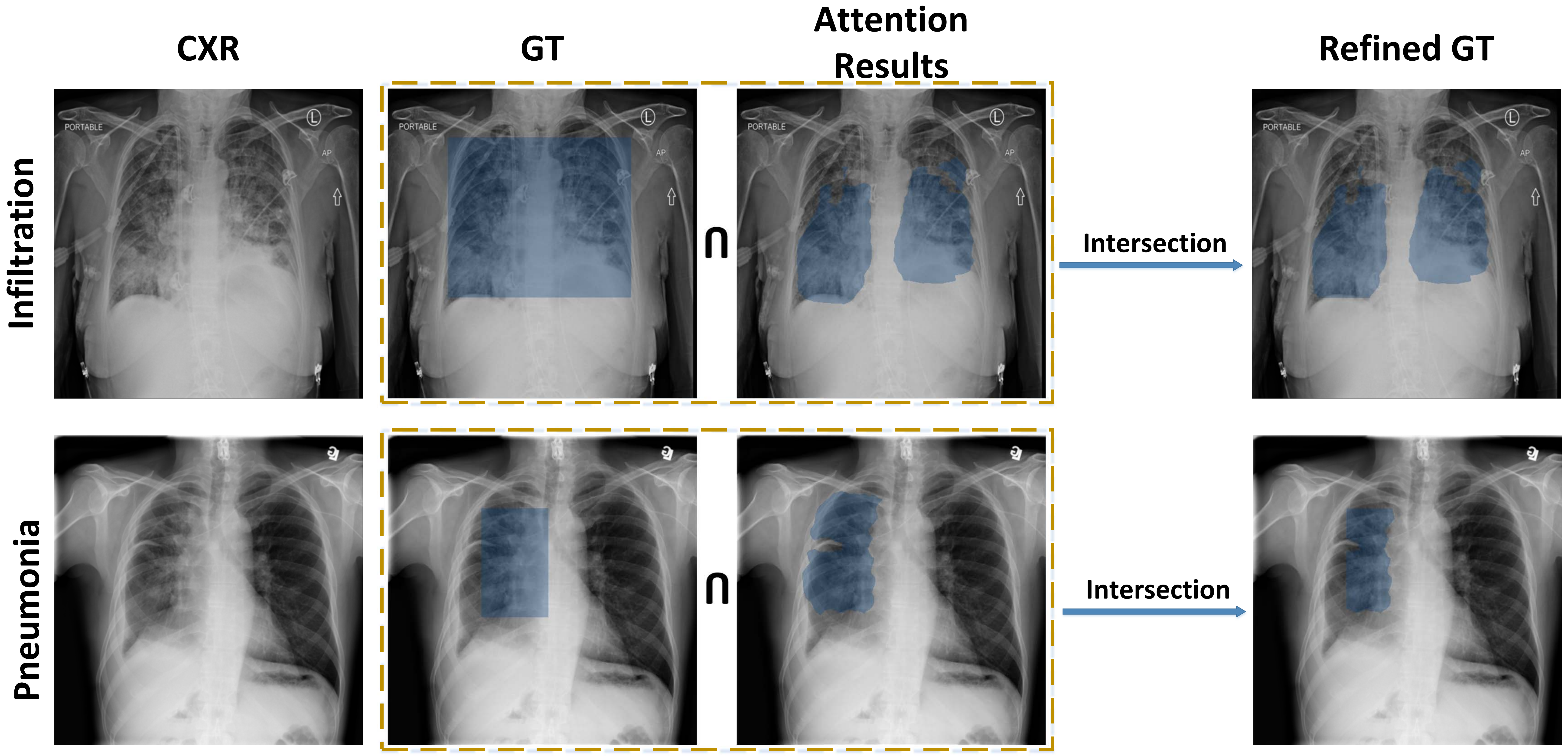}
\caption{
Illustration of the self-refinement method. We take the intersection of the box masks and attention results as the new refined ground-truth for the follow training.
}
\label{fig:sr-method}
\end{figure}

In real cases, although box annotations are helpful for the localization of abnormalities, the annotated boxes tend to be too large and enclose too much background cues. To address this issue, we further introduce a self-refinement (SR) method for the box annotations with extra pass of training. Specifically, we train an additional instance of our network with box annotations. Then, this additional network is used to generate attention maps for training images with box annotations for the potential refinement of boxes. If the IoU value of the attention region and the corresponding box is lower than 0.3, we keep the original box. 
Otherwise, we get the refined mask annotations via the overlap of the box masks and the corresponding attention maps. We show some examples in Fig. \ref{fig:sr-method}.
It is worth noting that the additional SR and main networks are distinct. The additional SR network is used for box refinement, whereas the main network is employed for the tasks of abnormality localization and diagnosis. The self-refinement method is not carried out on the testing data in the experiments.

\subsection{Overall Objective Loss Function}
Our overall learning objective is comprised of the abnormality and positive/negative classification, the attention bound and union, and the attention-adaptive MSE losses, and is expressed as: 
\begin{equation}
\label{eq:total}
{L_{total}} = {L_{ab}} + z \cdot {L_{amse}} + \lambda_{1}{L_{pn}} + \lambda_{2}{L_{bound}} + \lambda_{3}{L_{union}},
\end{equation}
where $L_{pn}$ and $L_{ab}$ are binary cross entropy (BCE) terms for the positive/negative and abnormality classifications, and $z$, $\lambda_{1}$, $\lambda_{2}$, and $\lambda_{3}$ are weighting factors (in our experiments we determine them with a search on $20\%$ of the training and validation set). Note that when box annotations are not used, we set $z=0$.




\section{Experiments}
\subsection{Datasets and Settings} 

\subsubsection{Experimental settings} 
We conduct experiments on the NIH Chest-Xray14 dataset \cite{wang2017chestx} and CheXpert dataset \cite{irvin2019chexpert}, two large-scale CXR datasets.
In our experiments, we use dilated ResNet50 \cite{yu2017dilated} as the backbone. The size of the abstracted feature map to generate the various attention maps is $\frac{1}{8}$ of the input images. We use the Adam \cite{kingma2014adam} optimizer with momentum set to $0.9$, a weight decay of $0.0001$, and a learning rate of $0.00002$ that is reduced by a factor of 10 after every $10$ epochs. We train our model with a batch size of 24 and empirically set $r=6$ for the LSE pooling, as defined in \cite{sun2016pronet}. During training, we perform random translation and horizontal flipping for augmentation. We resize the original 3-channel images to $512 \times 512$ as the input. The loss weight factors $z$, $\lambda_{1}$, $\lambda_{2}$, and $\lambda_{3}$ are set to 0.5, 0.01, 0.001, 0.001, respectively. 

\begin{table}[!t]
\newcommand{\tabincell}[2]{\begin{tabular}{@{}#1@{}}#2\end{tabular}}
\renewcommand{\arraystretch}{0.7}
\caption{Comparison of AUC scores with existing state-of-the-art methods on the official split of NIH Chest X-ray14. We report the AUC with 95\% confidence interval (CI) of our method.}
\label{table:auc}
\centering
\scalebox{0.78}{
\begin{tabular}{c|c|c|c|c|c}
\hline
&  &  &  &\\[-6pt]
Abnormality & \tabincell{c}{Wang \etal \\ \cite{wang2017chestx}} & \tabincell{c}{Li \etal \\ \cite{li2018thoracic}} & \tabincell{c}{DNetLoc \\ \cite{guendel2018learning}} & \tabincell{c}{CAN \\ \cite{ma2019cross}} & Ours \\
\hline
\hline
&  &  &  &\\[-5pt]
Atelectasis & 0.70 & 0.73 & 0.77 & \textbf{0.78} & 0.77 (0.77, 0.78) \\
\hline
&  &  &  &\\[-5pt]
Cardiomegaly & 0.81 & 0.84 & 0.88 & \textbf{0.89} & 0.87 (0.86, 0.88) \\
\hline
&  &  &  &\\[-5pt]
Effusion & 0.76 & 0.79 & \textbf{0.83} & \textbf{0.83} & \textbf{0.83} (0.83, 0.84) \\
\hline
&  &  &  &\\[-5pt]
Infiltration & 0.66 & 0.67 & \textbf{0.71} & 0.70 & \textbf{0.71} (0.71, 0.72) \\
\hline
&  &  &  &\\[-5pt]
Mass & 0.69 & 0.78 & 0.82 & \textbf{0.84} &  0.83 (0.82, 0.84) \\
\hline
&  &  &  &\\[-5pt]
Nodule & 0.67 & 0.70 & 0.76 & 0.77 & \textbf{0.79} (0.78, 0.81) \\
\hline
&  &  &  &\\[-5pt]
Pneumonia & 0.66 & 0.65 & \textbf{0.73} & 0.72 & 0.72 (0.70, 0.75) \\
\hline
&  &  &  &\\[-5pt]
Pneumothorax & 0.80 & 0.81 & 0.85 & 0.86 & \textbf{0.88} (0.87, 0.88) \\
\hline
&  &  &  &\\[-5pt]
Consolidation & 0.70 & 0.71 & \textbf{0.75} & \textbf{0.75} & 0.74 (0.73, 0.75) \\
\hline
&  &  &  &\\[-5pt]
Edema & 0.81 & 0.81 & 0.84 & \textbf{0.85} & 0.84 (0.83, 0.85) \\
\hline
&  &  &  &\\[-5pt]
Emphysema & 0.83 & 0.88 & 0.90 & 0.91 & \textbf{0.94} (0.93, 0.95) \\
\hline
&  &  &  &\\[-5pt]
Fibrosis & 0.79 & 0.77 & 0.82 & \textbf{0.83} & \textbf{0.83} (0.81, 0.85) \\
\hline
&  &  &  &\\[-5pt]
Pleural Thickening & 0.68 & 0.73 & 0.76 & \textbf{0.79} & \textbf{0.79} (0.78, 0.80) \\
\hline
&  &  &  &\\[-5pt]
Hernia & 0.87 & 0.69 & 0.90 & \textbf{0.93} & 0.91 (0.87, 0.94) \\
\hline
\hline
&  &  &  &\\[-5pt]
Mean AUC & 0.745 & 0.755 & 0.807 & 0.817 & \textbf{0.819} (0.815, 0.823)  \\
\hline
\end{tabular}
}
\end{table}

\begin{table}[!t]
\newcommand{\tabincell}[2]{\begin{tabular}{@{}#1@{}}#2\end{tabular}}
\renewcommand{\arraystretch}{0.7}
\caption{Comparison of AUC scores in 5-fold CV scheme of NIH Chest X-ray14. We show the standard deviation of Li \protect \etal \cite{li2018thoracic} and ours.}
\label{table:auc:5fold}
\centering
\scalebox{0.9}{
\begin{tabular}{c|c|c|c}
\hline
&  &  &  \\[-5pt]
Abnormality & \tabincell{c}{Li \etal \cite{li2018thoracic}} & \tabincell{c}{Liu \etal \cite{liu2019align}} & Ours \\
\hline
\hline
&  &  &  \\[-5pt]
Atelectasis &  0.80 $\pm$ 0.00 & 0.79 & \textbf{0.82} $\pm$ 0.01 \\
\hline
&  &  &  \\[-5pt]
Cardiomegaly & 0.87 $\pm$ 0.01 & 0.87 & \textbf{0.90} $\pm$ 0.02 \\
\hline
&  &  &  \\[-5pt]
Effusion & 0.87 $\pm$ 0.00 & \textbf{0.88} & \textbf{0.88} $\pm$ 0.01 \\
\hline
&  &  &  \\[-5pt]
Infiltration & 0.70 $\pm$ 0.01 & 0.69 & \textbf{0.72} $\pm$ 0.01 \\
\hline
&  &  &  \\[-5pt]
Mass & 0.83 $\pm$ 0.01 & 0.81 & \textbf{0.85} $\pm$ 0.02 \\
\hline
&  &  &  \\[-5pt]
Nodule & 0.75 $\pm$ 0.01 & 0.73 & \textbf{0.79} $\pm$ 0.01 \\
\hline
&  &  &  \\[-5pt]
Pneumonia & 0.67 $\pm$ 0.01 & \textbf{0.75} & 0.73 $\pm$ 0.01 \\
\hline
&  &  &  \\[-5pt]
Pneumothorax & 0.87 $\pm$ 0.01 & 0.89 & \textbf{0.90} $\pm$ 0.01 \\
\hline
&  &  &  \\[-5pt]
Consolidation & \textbf{0.80} $\pm$ 0.01 & 0.79 & \textbf{0.80} $\pm$ 0.01 \\
\hline
&  &  &  \\[-5pt]
Edema & 0.88 $\pm$ 0.01 & \textbf{0.91} & 0.90 $\pm$ 0.01 \\
\hline
&  &  &  \\[-5pt]
Emphysema & 0.91 $\pm$ 0.01 & 0.93 & \textbf{0.94} $\pm$ 0.01 \\
\hline
&  &  & \\[-5pt]
Fibrosis & 0.78 $\pm$ 0.02 & 0.80 & \textbf{0.81} $\pm$ 0.01 \\
\hline
&  &  &  \\[-5pt]
Pleural Thickening & 0.79 $\pm$ 0.01 & \textbf{0.80} & 0.79 $\pm$ 0.01  \\
\hline
&  &  &  \\[-5pt]
Hernia & 0.77 $\pm$ 0.03 & \textbf{0.92} & 0.86 $\pm$ 0.03  \\
\hline
\hline
&  &  &  \\[-5pt]
Mean AUC & 0.806 & 0.826 & \textbf{0.835} $\pm$ 0.007  \\
\hline
\end{tabular}
}
\end{table}

\begin{table*}[!t]
\newcommand{\tabincell}[2]{\begin{tabular}{@{}#1@{}}#2\end{tabular}}
\renewcommand{\arraystretch}{0.9}
\caption{Comparison of localization results trained using $100\%$ unannotated images of NIH Chest X-ray14.}
\label{table:iou:unannotated}
\centering
\scalebox{0.9}{
\begin{tabular}{c|c|c|c|c|c|c|c|c|c|c}
\hline
 $T(IoU)$ & Model & Atelectasis & Cardiomegaly & Effusion & Infiltration & Mass & Nodule & Pneumonia & Pneumothorax & Mean \\
\hline
\hline
&  &  &  & &  &  &  & & &\\[-7.5pt]
\multirow{3}{*}{0.1} & Li \etal \cite{li2018thoracic} & 0.59 & 0.81 & 0.73 & \textbf{0.85} & 0.69 & 0.29 & 0.23 & 0.38 & 0.57 \\
\cline{2-11}
&  &  &  & &  &  &  & & &\\[-7.5pt]
 & Liu \etal \cite{liu2019align} & 0.39 & 0.90 & 0.65 & \textbf{0.85} & 0.69 & 0.38 & 0.30 & 0.39 & 0.60 \\
\cline{2-11}
&  &  &  & &  &  &  & & &\\[-7.5pt]
& Ours & \textbf{0.78} & \textbf{0.97} & \textbf{0.82} & \textbf{0.85} & \textbf{0.78} & \textbf{0.56} & \textbf{0.76} & \textbf{0.48} & \textbf{0.75} \\
\hline
\hline
&  &  &  & &  &  &  & & &\\[-7.5pt]
\multirow{2}{*}{0.3} & Liu \protect\etal \cite{liu2019align} & \textbf{0.34} & \textbf{0.71} & \textbf{0.39} & \textbf{0.65} & 0.48 & 0.09 & 0.16 & 0.20 & \textbf{0.38}\\
\cline{2-11}
&  &  &  & &  &  &  & & &\\[-7.5pt]
& Ours & \textbf{0.34} & 0.40 & 0.27 & 0.55 & \textbf{0.51} & \textbf{0.14} & \textbf{0.42} & \textbf{0.22} & 0.36 \\
\hline
\end{tabular}
}
\end{table*}

\subsubsection{Evaluation metrics} 
We use the area under the ROC curve (AUC) to measure classification performance, and intersection over union ratio (IoU) and the intersection over the detected region (IoR) to quantify localization results. To calculate IoU and IoR, we use bounding boxes of localized attentive regions. Following prior work \cite{wang2017chestx,li2018thoracic}, we report the ratio of the number of cases with correct localization against the total number of cases in each class. A localization result is considered correct if the criterion of either IoU $> T(IoU)$ or IoR $> T(IoR)$, where $T(*)$ is the threshold, is met.

\subsection{NIH Chest-Xray14 Dataset}
NIH Chest-Xray14 dataset provides 112,120 X-ray images with abnormality labels from 30,805 patients. Images are labeled with 14 abnormality classes, with 984 bounding boxes of 8 abnormalities for 880 images labeled by board-certified radiologists. 

\subsubsection{Abnormality classification}
We conduct two experiments to evaluate the performance on the abnormality classification task and compare to the state-of-the-art (SOTA) methods. The first experiment is based on the official split \cite{wang2017chestx} and prepared at the patient-level. All images with box annotations are in the testing set and not used during model training. 
Results are shown in Table \ref{table:auc}, where our method outperforms state-of-the-art methods in terms of AUC. 
In particular, our method outperforms DNetLoc \cite{guendel2018learning} and CAN \cite{ma2019cross} methods that employ deeper models (DenseNet121) as their backbones. 

\begin{figure}[!t]
\centering
\includegraphics[width=8.4cm, height=10.8cm]{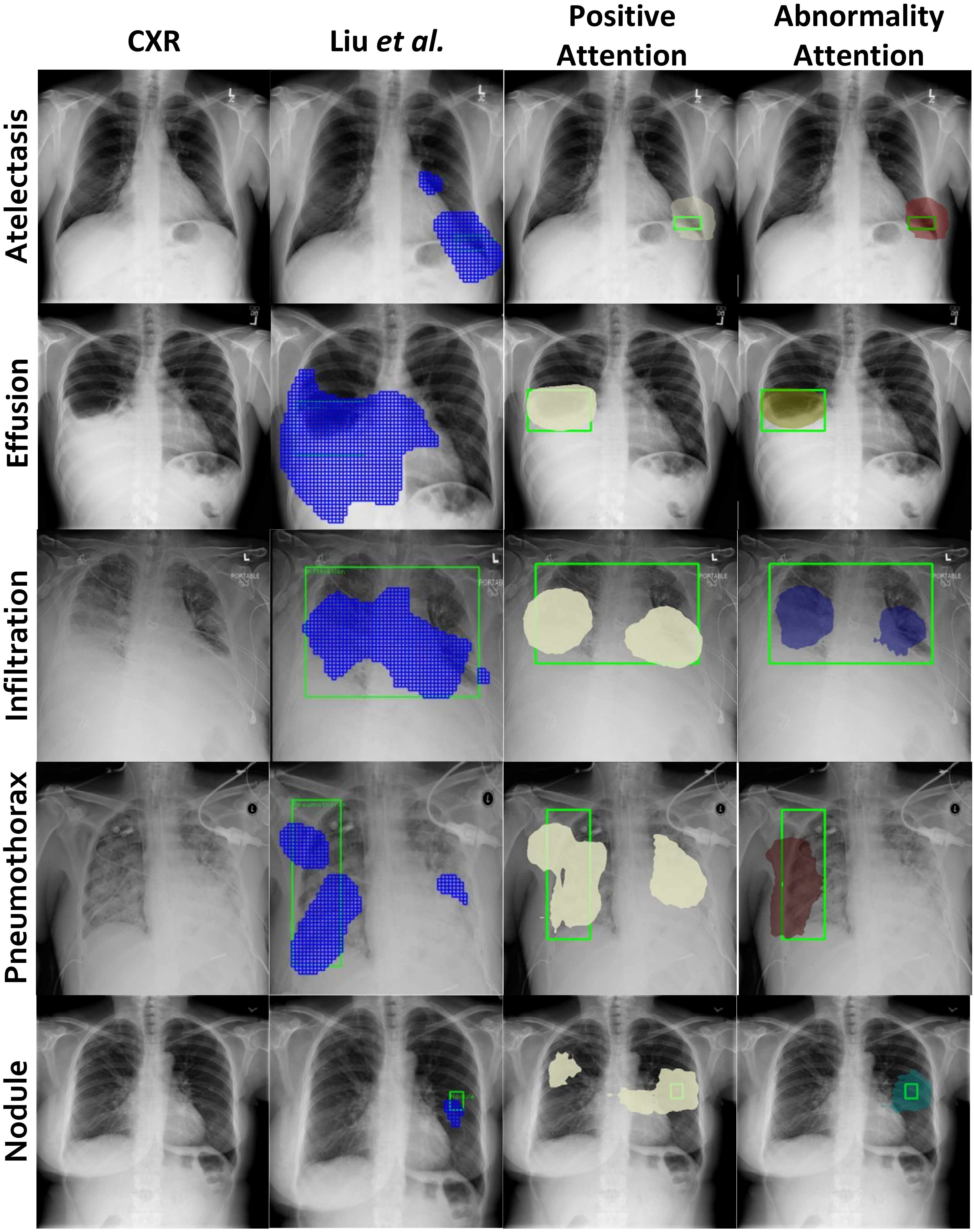}
\caption{Comparison of qualitative results between Liu \protect\etal \cite{liu2019align} and ours. All results are from the model trained with $100\%$ images without any box annotations. For the second column, Liu \protect\etal \cite{liu2019align} zoom in images for closer observation.}
\label{fig:comparison-iccv}
\end{figure}

In the second experiment, we use the 5-fold cross-validation (CV) scheme following \cite{li2018thoracic,liu2019align}. For the convenience of comparison, we use the same notations with \cite{li2018thoracic,liu2019align} by referring to unannotated CXR images as those with only image-level labels and annotated CXR images as those with both image- and box-level labels.  In each fold, we use $70\%$ of the annotated and $70\%$ of unannotated images for training, 
and $10\%$ of the annotated and unannotated images for validation. Then, the rest $20\%$ of the annotated and unannotated images are used for testing. We summarize AUC scores of the compared methods \normalem\emph{w.r.t.} the 14 abnormalities of testing set in Table \ref{table:auc:5fold}, where our method obtains the best mean AUC score of $0.835$.

\begin{table*}[!t]
\newcommand{\tabincell}[2]{\begin{tabular}{@{}#1@{}}#2\end{tabular}}
\renewcommand{\arraystretch}{0.8}
\caption{Results of models trained with $50\%$ unannotated and $80\%$ annotated images of NIH Chest X-ray14 at various $T(IoU)$ thresholds.}
\label{table:iou}
\centering
\scalebox{0.9}{
\begin{tabular}{c|c|c|c|c|c|c|c|c|c|c}
\hline
&  &  &  &  &  & & & & & \\[-7.0pt]
$T(IoU)$ & Model & Atelectasis & Cardiomegaly & Effusion & Infiltration & Mass & Nodule & Pneumonia & Pneumothorax & Mean \\
\hline
\hline
&  &  &  &  &  & & & & & \\[-6.5pt]
\multirow{3}{*}{0.1} & Wang \protect\etal \cite{wang2017chestx} & 0.69 & 0.94 & 0.66 & 0.71 & 0.40 & 0.14 & 0.63 & 0.38 & 0.57\\
\cline{2-11}
&  &  &  &  &  & & & & & \\[-6.0pt]
 & Li \protect\etal \cite{li2018thoracic} & \textbf{0.71} & 0.98 & 0.87 & \textbf{0.92} & 0.71 & 0.40 & 0.60 & 0.63 & 0.73\\
\cline{2-11}
&  &  &  &  &  & & & & & \\[-6.0pt]
 & Ours & \textbf{0.71} & \textbf{1.00} & \textbf{0.89} & 0.88 & \textbf{0.76} & \textbf{0.65} & \textbf{0.91} & \textbf{0.78} & \textbf{0.82} \\
\hline
\hline
&  &  &  &  &  & & & & & \\[-6.5pt]
\multirow{3}{*}{0.2} & Wang \protect\etal \cite{wang2017chestx} & 0.47 & 0.68 & 0.45 & 0.48 & 0.26 & 0.05 & 0.35 & 0.23 & 0.37\\
\cline{2-11}
&  &  &  &  &  & & & & & \\[-6.0pt]
 & Li \protect\etal \cite{li2018thoracic} & 0.53 & 0.97 & \textbf{0.76} & \textbf{0.83} & 0.59 & 0.29 & 0.50 & 0.51 & 0.62\\
\cline{2-11}
&  &  &  &  &  & & & & & \\[-6.0pt]
 & Ours & \textbf{0.54} & \textbf{1.00} & 0.75 & 0.79 & \textbf{0.67} & \textbf{0.53} & \textbf{0.86} & \textbf{0.60} & \textbf{0.72} \\
\hline
\hline
&  &  &  &  &  & & & & & \\[-6.5pt]
\multirow{4}{*}{0.3} & Wang \protect\etal \cite{wang2017chestx} & 0.24 & 0.46 & 0.30 & 0.28 & 0.15 & 0.04 & 0.16 & 0.13 & 0.22 \\
\cline{2-11}
&  &  &  &  &  & & & & & \\[-6.0pt]
 & Li \protect\etal \cite{li2018thoracic} & 0.36 & 0.94 & 0.56 & 0.66 & 0.45 & 0.17 & 0.39 & \textbf{0.44} & 0.49 \\
 \cline{2-11}
 &  &  &  &  &  & & & & & \\[-6.0pt]
 & Liu \protect\etal \cite{liu2019align} & \textbf{0.53} & 0.88 & \textbf{0.57} & \textbf{0.73} & 0.48 & 0.10 & 0.49 & 0.40 & 0.53 \\
\cline{2-11}
&  &  &  &  &  & & & & & \\[-6.0pt]
 & Ours & 0.40 & \textbf{1.00} & 0.52 & 0.68 & \textbf{0.58} & \textbf{0.46} & \textbf{0.69} & 0.43 & \textbf{0.60} \\
\hline
\hline
&  &  &  &  &  & & & & & \\[-6.5pt]
\multirow{3}{*}{0.4} & Wang \protect\etal \cite{wang2017chestx} & 0.09 & 0.28 & 0.20 & 0.12 & 0.07 & 0.01 & 0.08 & 0.07 & 0.12 \\
\cline{2-11}
&  &  &  &  &  & & & & & \\[-6.0pt]
 & Li \protect\etal \cite{li2018thoracic} & \textbf{0.25} & 0.88 & \textbf{0.37} & 0.50 & 0.33 & 0.11 & 0.26 & 0.29 & 0.42 \\
\cline{2-11}
&  &  &  &  &  & & & & & \\[-6.0pt]
 & Ours & 0.26 & \textbf{1.00} & 0.29 & \textbf{0.56} & \textbf{0.40} & \textbf{0.35} & \textbf{0.50} & \textbf{0.32} & \textbf{0.46} \\
\hline
\hline
&  &  &  &  &  & & & & & \\[-6.5pt]
\multirow{4}{*}{0.5} & Wang \protect\etal \cite{wang2017chestx} & 0.05 & 0.18 & 0.11 & 0.07 & 0.01 & 0.01 & 0.03 & 0.03 & 0.06 \\
\cline{2-11}
&  &  &  &  &  & & & & & \\[-6.0pt]
 & Li \protect\etal \cite{li2018thoracic} & 0.14 & 0.84 & 0.22 & 0.30 & 0.22 & 0.07 & 0.17 & 0.19 & 0.27 \\
 \cline{2-11}
 &  &  &  &  &  & & & & & \\[-6.0pt]
& Liu \protect\etal \cite{liu2019align} & \textbf{0.32} & 0.78 & \textbf{0.40} & \textbf{0.61} & \textbf{0.33} & 0.05 & \textbf{0.37} & \textbf{0.23} & \textbf{0.39} \\
\cline{2-11}
&  &  &  &  &  & & & & & \\[-6.0pt]
 & Ours & 0.15 & \textbf{0.99} & 0.14 & 0.33 & 0.27 & \textbf{0.22} & 0.35 & 0.22 & 0.33 \\
\hline
\hline
&  &  &  &  &  & & & & & \\[-6.5pt]
\multirow{3}{*}{0.6} & Wang \protect\etal \cite{wang2017chestx} & 0.02 & 0.08 & 0.05 & 0.02 & 0.00 & 0.01 & 0.02 & 0.03 & 0.03 \\
\cline{2-11}
&  &  &  &  &  & & & & & \\[-6.0pt]
& Li \protect\etal \cite{li2018thoracic} & 0.07 & 0.73 & \textbf{0.15} & \textbf{0.18} & \textbf{0.16} & 0.03 & 0.10 & \textbf{0.12} & 0.19 \\
\cline{2-11}
&  &  &  &  &  & & & & & \\[-6.0pt]
 & Ours & \textbf{0.08} & \textbf{0.97} & 0.05 & \textbf{0.18} & 0.14 & \textbf{0.15} & \textbf{0.27} & 0.11 & \textbf{0.24} \\
\hline
\hline
&  &  &  &  &  & & & & & \\[-6.5pt]
\multirow{4}{*}{0.7} & Wang \protect\etal \cite{wang2017chestx} & 0.01 & 0.03 & 0.02 & 0.00 & 0.00 & 0.00 & 0.01 & 0.02 & 0.01 \\
\cline{2-11}
&  &  &  &  &  & & & & & \\[-6.0pt]
 & Li \protect\etal \cite{li2018thoracic} & 0.04 & 0.52 & 0.07 & 0.09 & 0.11 & 0.01 & 0.05 & 0.05 & 0.12 \\
 \cline{2-11}
 &  &  &  &  &  & & & & & \\[-6.0pt]
& Liu \protect\etal \cite{liu2019align} & \textbf{0.18} & 0.70 & \textbf{0.28} & \textbf{0.41} & \textbf{0.27} & 0.04 & \textbf{0.25} & \textbf{0.18} & \textbf{0.29} \\
\cline{2-11}
&  &  &  &  &  & & & & & \\[-6.0pt]
 & Ours & 0.02 & \textbf{0.77} & 0.01 & 0.12 & 0.08 & \textbf{0.10} & 0.06 & 0.03 & 0.15 \\
\hline
\end{tabular}
}
\end{table*}

\begin{table*}[!ht]
\newcommand{\tabincell}[2]{\begin{tabular}{@{}#1@{}}#2\end{tabular}}
\renewcommand{\arraystretch}{0.8}
\caption{Results of models trained with $50\%$ unannotated and $80\%$ annotated images of NIH Chest X-ray14 at various $T(IoR)$ thresholds. Liu \protect\etal \cite{liu2019align} does not list any $T(IoR)$ results in their paper. ``SR" indicates the proposed self-refinement method to reduce the noise of box annotations.}
\label{table:ior}
\centering
\scalebox{0.9}{
\begin{tabular}{c|c|c|c|c|c|c|c|c|c|c}
\hline
&  &  &  &  &  & & & & & \\[-7.0pt]
$T(IoR)$ & Model & Atelectasis & Cardiomegaly & Effusion & Infiltration & Mass & Nodule & Pneumonia & Pneumothorax & Mean \\
\hline
\hline
&  &  &  &  &  & & & & & \\[-6.5pt]
\multirow{4}{*}{0.1} & Wang \protect\etal \cite{wang2017chestx} & 0.62 & \textbf{1.00} & 0.80 & 0.91 & 0.59 & 0.15 & 0.86 & 0.52 & 0.68\\
\cline{2-11}
&  &  &  &  &  & & & & & \\[-6.0pt]
 & Li \protect\etal \cite{li2018thoracic} & \textbf{0.77} & 0.99 & 0.91 & \textbf{0.95} & 0.75 & 0.40 & 0.69 & 0.68 & 0.77\\
\cline{2-11}
&  &  &  &  &  & & & & & \\[-6.0pt]
 & Ours & 0.74 & \textbf{1.00} & 0.92 & 0.88 & 0.81 & 0.67 & \textbf{0.94} & \textbf{0.80} & 0.85 \\
 \cline{2-11}
&  &  &  &  &  & & & & & \\[-6.0pt]
 & Ours + SR & 0.69 & \textbf{1.00} & \textbf{0.93} & 0.91 & \textbf{0.84} & \textbf{0.71} & \textbf{0.94} & 0.79 & \textbf{0.85} \\
\hline
\hline
&  &  &  &  &  & & & & & \\[-6.5pt]
\multirow{4}{*}{0.25} & Wang \protect\etal \cite{wang2017chestx} & 0.39 & 0.99 & 0.63 & 0.80 & 0.46 & 0.05 & 0.71 & 0.34 & 0.55\\
\cline{2-11}
&  &  &  &  &  & & & & & \\[-6.0pt]
 & Li \protect\etal \cite{li2018thoracic} & 0.57 & 0.99 & 0.79 & \textbf{0.88} & 0.57 & 0.25 & 0.62 & 0.61 & 0.66\\
\cline{2-11}
&  &  &  &  &  & & & & & \\[-6.0pt]
 & Ours & \textbf{0.61} & \textbf{1.00} & 0.76 & 0.82 & 0.74 & 0.54 & \textbf{0.89} & 0.62 & 0.75 \\
 \cline{2-11}
&  &  &  &  &  & & & & & \\[-6.0pt]
 & Ours + SR & 0.60 & \textbf{1.00} & \textbf{0.84} & 0.87 & \textbf{0.76} & \textbf{0.59} & 0.88 & \textbf{0.72} & \textbf{0.78} \\
\hline
\hline
&  &  &  &  &  & & & & & \\[-6.5pt]
\multirow{4}{*}{0.5} & Wang \protect\etal \cite{wang2017chestx} & 0.19 & 0.95 & 0.42 & 0.65 & 0.31 & 0.00 & 0.48 & 0.27 & 0.41 \\
\cline{2-11}
&  &  &  &  &  & & & & & \\[-6.0pt]
 & Li \protect\etal \cite{li2018thoracic} & 0.35 & 0.98 & 0.52 & 0.62 & 0.40 & 0.11 & 0.49 & 0.43 & 0.49 \\
\cline{2-11}
&  &  &  &  &  & & & & & \\[-6.0pt]
 & Ours & 0.40 & \textbf{1.00} & 0.50 & 0.59 & 0.60 & 0.35 & 0.63 & 0.43 & 0.56 \\
 \cline{2-11}
&  &  &  &  &  & & & & & \\[-6.0pt]
 & Ours + SR & \textbf{0.46} & \textbf{1.00} & \textbf{0.61} & \textbf{0.72} & \textbf{0.61} & \textbf{0.38} & \textbf{0.72} & \textbf{0.57} & \textbf{0.63} \\
\hline
\hline
&  &  &  &  &  & & & & & \\[-6.5pt]
\multirow{4}{*}{0.75} & Wang \protect\etal \cite{wang2017chestx} & 0.09 & 0.82 & 0.23 & 0.44 & 0.16 & 0.00 & 0.29 & 0.17 & 0.28 \\
\cline{2-11}
&  &  &  &  &  & & & & & \\[-6.0pt]
 & Li \protect\etal \cite{li2018thoracic} & 0.20 & 0.87 & 0.34 & 0.46 & 0.29 & 0.07 & 0.43 & 0.30 & 0.37 \\
\cline{2-11}
&  &  &  &  &  & & & & & \\[-6.0pt]
 & Ours & 0.21 & 0.88 & 0.28 & 0.38 & 0.36 & \textbf{0.25} & 0.43 & 0.27 & 0.38 \\
 \cline{2-11}
&  &  &  &  &  & & & & & \\[-6.0pt]
 & Ours + SR & \textbf{0.25} & \textbf{0.97} & \textbf{0.41} & \textbf{0.49} & \textbf{0.51} & \textbf{0.25} & \textbf{0.45} & \textbf{0.37} & \textbf{0.46} \\
\hline
\hline
&  &  &  &  &  & & & & & \\[-6.5pt]
\multirow{4}{*}{0.9} & Wang \protect\etal \cite{wang2017chestx}  & 0.07 & 0.65 & 0.14 & \textbf{0.36} & 0.09 & 0.00 & 0.23 & 0.12 & 0.21 \\
\cline{2-11}
&  &  &  &  &  & & & & & \\[-6.0pt]
 & Li \protect\etal \cite{li2018thoracic} & \textbf{0.15} & 0.59 & 0.23 & 0.32 & 0.22 & 0.06 & \textbf{0.34} & 0.22 & 0.27 \\
\cline{2-11}
&  &  &  &  &  & & & & & \\[-6.0pt]
 & Ours & 0.10 & 0.58 & 0.11 & 0.20 & 0.18 & 0.16 & 0.23 & 0.13 & 0.21 \\
 \cline{2-11}
&  &  &  &  &  & & & & & \\[-6.0pt]
 & Ours + SR & 0.12 & \textbf{0.75} & \textbf{0.25} & 0.27 & \textbf{0.28} & \textbf{0.20} & 0.28 & \textbf{0.28} & \textbf{0.30} \\
\hline
\end{tabular}
}
\end{table*}


\subsubsection{Abnormality localization}
\label{section:localizaition}
Two experiments for assessing localization performance are conducted. 
In the first experiment, we show localization performance by considering only image-level annotations, \ie, abnormality labels. Following \cite{li2018thoracic,liu2019align}, we train our model with $100\%$ images (111,240) without any box annotations and test with the 880 images with box annotations. We summarize our results in Table~\ref{table:iou:unannotated}, where we compare to SOTA methods at $T(IoU)=0.1$ and $0.3$. Our method significantly improves localization performance for all abnormalities at $T(IoU)=0.1$, with a mean score of $0.75$. In particular, we achieve more than $150\%$ improvement in localization for ``Pneumonia" compared to the results of the most recent SOTA method \cite{liu2019align}. We also observe remarkable localization improvements for the ``Atelectasis", ``Mass" and ``Nodule" abnormalities. In $T(IoU)=0.3$, our method also yields competitive localization performance and outperforms \cite{liu2019align} for ``Mass", ``Nodule" and ``Pneumonia". We further show qualitative comparison of the attention maps with SOTA method \cite{liu2019align} in Fig. \ref{fig:comparison-iccv}, where the exactness of our attention map can be corroborated.

Also, we notice that our proposed method does not perform well for ``Cardiomegaly", ``Effusion" and ``Infiltration" at $T(IoU)=0.3$ compared with Liu \protect\etal \cite{liu2019align}. The underlying reasons may be twofold: 1) the threshold setting for the attention maps; 2) the coverage range of bounding boxes. For the first reason, we set a very high threshold (0.999) on attention maps to get the final binary localization masks for the computation of evaluation scores in Tables \ref{table:iou:unannotated}. This threshold is very rigorous and may generate relatively small object masks, which may not be favorable for large class like ``Cardiomegaly", for localization. For comparison, we further explore a smaller attention threshold of 0.1. The resulting correct ratio score at $T(IoU) = 0.3$ with the attention threshold of 0.1 for ``Cardiomegaly" is 0.72, which is slightly higher than result of Liu \etal \cite{liu2019align}. The second reason is the issue of box definition that may lead to lower performance of our method at $T(IoU) = 0.3$, particularly for ``Effusion" and ``Infiltration". We have invited an experienced radiologist to carefully review the original annotated boxes and suggest that the variety of the annotation coverage for some classes (\eg, ``Effusion", ``Infiltration" and ``Pneumonia") is very large (see figures and detailed analysis for better illustration in the supplementary material). 
Larger boxes with more non-related regions may favor localization results with relatively larger masks for higher scores. Specifically, the results of Liu \etal \cite{liu2019align} for the ``Infiltration" and ``Effusion" cases in the Fig. \ref{fig:comparison-iccv} are fuzzy and cover more non-related regions (even the abdominal region), while our localization results are closer to the abnormal regions of ``Effusion" and ``Infiltration". Therefore, the results of Liu \etal \cite{liu2019align} for ``Effusion" and ``Infiltration" may be higher.


In the second experiment, we use the 5-fold CV scheme following \cite{li2018thoracic,liu2019align}. In each fold, we train our model with $50\%$ of the unannotated images and $80\%$ of the annotated images, and tested with the remaining $20\%$ of the annotated images. 
As shown in Table \ref{table:iou}, our method outperforms SOTA methods in most cases. In particular, for the difficult classes of ``Mass" and ``Nodule", our localization performance is remarkably better. 
Similar with results in Table \ref{table:iou}, our method does not perform well for some abnormalities (\eg, ``Effusion" and ``Infiltration"). Like previous observation, it may be due to large box annotations for these abnormalities. For ``Pneumonia", we notice that our method achieves much better performance than baselines at $T(IoU)<0.5$ but much worse at $T(IoU)>=0.5$.
Although most boxes provided by NIH for ``Pneumonia" are reasonable, some box annotations also enclose many non-related regions. Meanwhile, it is worth noting that there are some missing boxes in NIH Chest X-ray14. The figures and more details can be found in the supplementary material. 
Thus, we suggest the results with IoU scores in the range of 0.3 to 0.7 are acceptable (see Fig. 9 and Fig. 10 in the supplementary material). It can be observed that nearly 63\% ($0.69-0.06$) of our results have IoU scores of ``Pneumonia" in the range of 0.3 to 0.7, compared to 24\% ($0.49-0.25$) of results in Liu \etal \cite{liu2019align}.

\begin{figure}[!t]
\centering
\includegraphics[width=7.0cm, height=8.8cm]{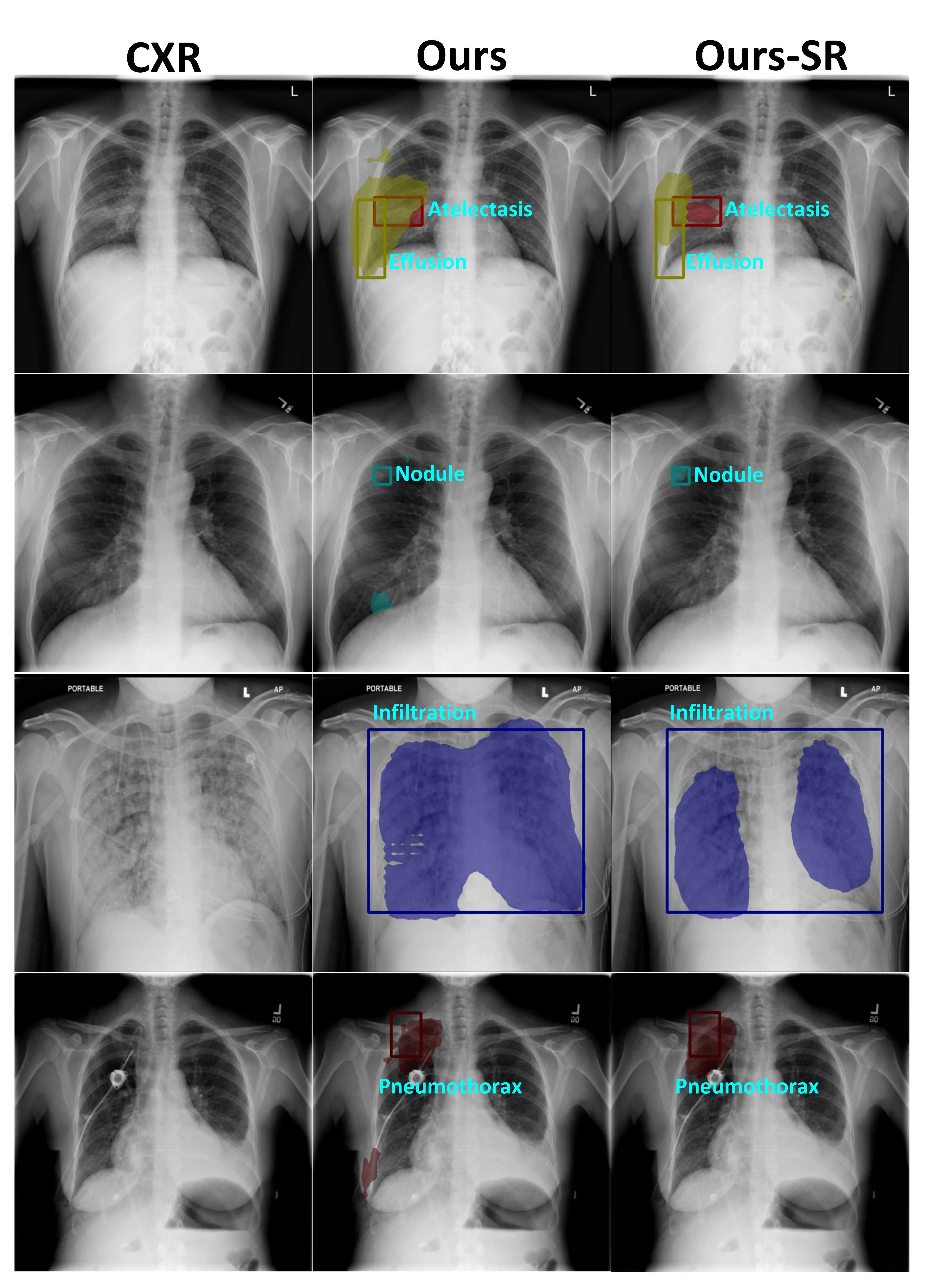}
\caption{Comparison of qualitative results using the self-refinement method. The images in the first row are input CXR images. The second row and third row show abnormality attention maps from the model trained with box annotations or the refined masks, respectively.}
\label{fig:sr}
\end{figure}

\begin{table*}[!t]
\setlength\tabcolsep{3pt}
\newcommand{\tabincell}[2]{\begin{tabular}{@{}#1@{}}#2\end{tabular}}
\renewcommand{\arraystretch}{0.7}
\caption{Ablation study of all component in our method (RN50 = ResNet50, PA = Positive Attention, ABU = Attention Bound and Union Loss). All models are trained with the official split of NIH Chest X-ray14.}
\label{table:ablation}
\centering
\scalebox{0.92}{
\begin{tabular}{l|c|c|c|c|c|c|c|c|c|c}
\hline
\multirow{2}{*}{Model} & \multirow{2}{*}{\tabincell{c}{Mean \\ AUC}} & \multicolumn{9}{c}{$T(IoU)=0.1$} \\
\cline{3-11}
& &  &  &  &  &  &  &  &  &\\[-5.5pt]
& & Atelectasis & Cardiomegaly & Effusion & Infiltration & Mass & Nodule & Pneumonia & Pneumothorax & Mean \\
\hline
\hline
& &  &  &  &  &  &  &  &  &\\[-5.5pt]
RN50-32-GAP & 0.816 & 0.36 & 0.65 & 0.60 & 0.53 & 0.38 & 0.04 & 0.01 & 0.38 & 0.37\\
 \hline
 & &  &  &  &  &  &  &  &  &\\[-5.5pt]
RN50-8-GAP & 0.815 & 0.33 & \ 0.61 & 0.61 & 0.35 & 0.48 & 0.19 & 0.13 & 0.43 & 0.39\\

\hline
& &  &  &  &  &  &  &  &  &\\[-5.5pt]
RN50-8-LSE & 0.816 & 0.38 & 0.58 & 0.42 & 0.24 & 0.55 & 0.48 & 0.32 & \textbf{0.55} & 0.44\\
\hline
\hline
& &  &  &  &  &  &  &  &  &\\[-5.5pt]
\tabincell{c}{RN50-8-LSE+FAB-C} & 0.817 & 0.58 & 0.53 & 0.56 & 0.55 & 0.59 & \textbf{0.59} & 0.41 & 0.30 & 0.51\\
\hline
& &  &  &  &  &  &  &  &  &\\[-5.5pt]
\tabincell{c}{RN50-8-LSE+FAB-P} & 0.815 & 0.61 & 0.55 & 0.62 & 0.67 & 0.60 & 0.54 & 0.50 & 0.50 & 0.57\\
 \hline
 & &  &  &  &  &  &  &  &  &\\[-5.5pt]
\tabincell{c}{RN50-8-LSE+FAB} & 0.818 & 0.64 & \textbf{0.92} & 0.69 & 0.69 & 0.64 & 0.54 &  0.44 & 0.35 & 0.61\\
  \hline
 & &  &  &  &  &  &  &  &  &\\[-5.5pt]
\tabincell{c}{RN50-8-LSE+FAB+PA} &  0.815 & 0.61 & 0.91 & 0.70 & 0.76 & 0.67 & 0.51 & 0.58  & 0.31 & 0.63\\
\hline
 & &  &  &  &  &  &  &  &  &\\[-5.5pt]
\tabincell{c}{RN50-8-LSE+PA+ABU} & 0.817 & 0.37 & 0.88 & 0.50 & 0.82 & 0.60 & 0.15 & \textbf{0.68} & 0.43 & 0.55\\
\hline
 & &  &  &  &  &  &  &  &  &\\[-5.5pt]
\tabincell{c}{RN50-8-LSE+FAB+PA+Bound} & 0.816 & 0.68 & 0.78 & 0.69 & 0.78 & 0.68 & 0.52 & 0.61 & 0.46 & 0.65\\
\hline
 & &  &  &  &  &  &  &  &  &\\[-5.5pt]
 \tabincell{c}{RN50-8-LSE+FAB+PA+ABU} & \textbf{0.819}& \textbf{0.70} & 0.87 & \textbf{0.75} & \textbf{0.89} & \textbf{0.69} & \textbf{0.59} & \textbf{0.68} & 0.40 & \textbf{0.70}\\
\hline
\end{tabular}
}
\end{table*}

\begin{table*}[!t]
\newcommand{\tabincell}[2]{\begin{tabular}{@{}#1@{}}#2\end{tabular}}
\renewcommand{\arraystretch}{0.7}
\caption{Effectiveness of AMSE loss. All models are trained using $50\%$ unannotated and $80\%$ annotated images of NIH Chest X-ray14.}
\label{table:amse}
\centering
\scalebox{0.9}{
\begin{tabular}{c|c|c|c|c|c|c|c|c|c|c}
\hline
& & &  &  &  &  &  &  &  & \\[-6.5pt]
 $T(IoU)$ & Model & Atelectasis & Cardiomegaly & Effusion & Infiltration & Mass & Nodule & Pneumonia & Pneumothorax & Mean \\
\hline
\hline
& & &  &  &  &  &  &  &  & \\[-5.5pt]
\multirow{2}{*}{0.1} & Ours & \textbf{0.71} & \textbf{1.00} & \textbf{0.89} & 0.88 & \textbf{0.76} & \textbf{0.65} & \textbf{0.91} & \textbf{0.78} & \textbf{0.82} \\
\cline{2-11}
& & &  &  &  &  &  &  &  & \\[-5.0pt]
& Ours-MSE & 0.59 & 1.00 & 0.83 & \textbf{0.92} & 0.75 & 0.54 & 0.78 & 0.65 & 0.76 \\
\hline
\hline
& & &  &  &  &  &  &  &  & \\[-5.5pt]
\multirow{2}{*}{0.3} & Ours & \textbf{0.40} & \textbf{1.00} & \textbf{0.52} & \textbf{0.68} & \textbf{0.58} & \textbf{0.46} & \textbf{0.69} & \textbf{0.43} & \textbf{0.60} \\
\cline{2-11}
& & &  &  &  &  &  &  &  & \\[-5.0pt]
& Ours-MSE & 0.33 & 1.00 & 0.41 & 0.58 & 0.52 & 0.34 & 0.58 & 0.39 & 0.52 \\
\hline
\end{tabular}
}
\vspace{-1em}
\end{table*}

To better illustrate the localization performance, we also show the $T(IoR)$ results in Table \ref{table:ior}. It calculates the intersection over the detected region, reducing the effect from non-related regions in the box annotations in evaluation. As shown in Table \ref{table:ior}, our method gives the best performance in most items of $T(IoR)$ values. At the same time, we show the performance of the self-refinement method which can reduce the noise of the foreground in the box annotations during training. It proves that the self-refinement method can make the attention results more concentrated within the box annotations. We show qualitative results of the self-refinement method in Fig. \ref{fig:sr}.

In summary, our proposed hierarchical attention framework can result in more precise and pathological plausible localization of abnormalities and achieves state-of-the-art results at several $T(IoU)$ and $T(IoR)$ settings in our experiments. It is worth noting that our method localizes the relatively small abnormalities like ``Nodule" much better. The detection of pulmonary ``Nodule" in CXR images is also a very challenging task with low sensitivity in the range of $0.69$ to $0.82$ for radiologists \cite{Nam2018Radiology}. 

\subsubsection{Ablation Study}
Table \ref{table:ablation} shows results of the ablation studies with the AUC metrics and localization results of $T(IoU)=0.1$, using the official split for training and testing. Abnormality attention is applied in all the models to generate the abnormality localization prediction, which is the same as CAM \cite{zhou2016learning} when not using our attention bound and union loss.


First, ResNet50 without the positive/negative classification branch is implemented with down-sampled input images (factor of $32$). The second backbone is the dilated ResNet50 with images down-sampled by a factor of $8$. The global average pooling (GAP) is utilized for the two backbones. The corresponding results are shown in the first two rows of ``RN50-32-GAP" and ``RN50-8-GAP", respectively, in the Table \ref{table:ablation}. We can see that higher resolution, \ie, smaller down-sampling factor, is helpful for the localization of classes like ``Mass" and ``Nodule".
Second, we propose to use the LSE pooling instead of GAP. To illustrate the effectiveness, we compare the network of ``RN50-8-LSE" to ``RN50-8-GAP". 
As shown in Table \ref{table:ablation}, the average localization score is higher with LSE. 

Third, we perform an ablation study for different versions of FAB. Specifically, we compare three attention configurations: FAB-C (channel-wise only), FAB-P (position-wise only), and FAB (both channel-wise and position-wise). 
The combination of both attention modules achieves the best average localization performance (0.61). 

Fourth, we add the positive/negative classification branch to generate the positive attention (``RN50-8-LSE+FAB+PA"), with which the average localization score (0.63) is slightly higher than ``RN50-8-LSE+FAB". However, with the addition of our attention bound and union loss in the last row, the average localization performance increases to 0.70. The results suggest that the proposed attention losses are important for the hierarchical attention representation in guiding the learning of abnormality attention from positive attention. Results of our method without FAB (``RN50-8-LSE+PA+ABU") suggest that the FAB module is efficient for refining the feature encoding from the backbone network. Comparing the results of ``RN50-8-LSE+FAB" and the last row, we can see that the localization results of the ``Infiltration" and ``Pneumonia" are significantly boosted by our hierarchical attention mining method. In particular, the ``Pneumonia" class has the smallest number of samples (876 images) in the training set (86,524 images) of the official split. It is evident that our HAM method can alleviate the data-imbalance problem even in such an extreme situation. 

At the same time, we show the results of our method with only attention bound constraint (``RN50-8-LSE+FAB+PA+Bound"). It performs worse in both classification and localization tasks than the model with attention bound and union losses. 
Because there is no constraint for positive attention without the incorporation of attention union loss, the incorrect prediction of positive attention may misdirect the abnormality attention. In such a case, the final classification and localization performances are compromised. Accordingly, by comparing the ablation of attention bound and attention union losses, the effectiveness of the synergy of two attention losses is corroborated.

To better illustrate the statistical significance of our method, we also calculate the $p$-value between the baseline ``RN50-8-LSE" and our model ``RN50-8-LSE+FAB+PA+ABU" in Table \ref{table:ablation}. The $p$-value for the classification predictions of these two models is $4.31 \times 10^{-5}$, implying that the proposed methods have significant improvements compared with ``RN50-8-LSE".

Finally, the effectiveness of the proposed AMSE loss is shown by comparing the performance to the original MSE loss. We conduct the same 5-fold CV experiment as the second experiment in section \ref{section:localizaition}.
As shown in Table \ref{table:amse}, the AMSE loss is especially effective for the smaller abnormalities (\protect\eg, ``Nodule").

\begin{table}[!t]
\newcommand{\tabincell}[2]{\begin{tabular}{@{}#1@{}}#2\end{tabular}}
\renewcommand{\arraystretch}{0.65}
\caption{Comparison of AUC scores with different models ((RN50 = ResNet50, RN152 = ResNet152) on CheXpert validation set.}
\label{table:auc:chexpert}
\centering
\scalebox{0.78}{
\begin{tabular}{c|c|c|c|c|c|c}
\hline
&  &  &  &  &  &\\[-5pt]
Model & Atelectasis & Cardiomegaly & Consolidation & Edema & \tabincell{c}{Pleural \\ Effusion} & \tabincell{c}{Mean \\ AUC}\\
\hline
\hline
&  &  &  &  &  &\\[-5.5pt]
\tabincell{c}{U-Ignore \\ \cite{irvin2019chexpert}} & 0.818 & 0.828 & \textbf{0.938} & 0.934 & 0.928 & 0.8892\\
\hline
&  &  &  &  &  &\\[-5.5pt]
\tabincell{c}{U-Zeros \\ \cite{irvin2019chexpert}} & 0.811 & 0.840 & 0.932 & 0.929 & 0.931 & 0.8886 \\
\hline
&  &  &  &  &  &\\[-5.5pt]
\tabincell{c}{U-Ones \\ \cite{irvin2019chexpert}} & 0.858 & 0.832 & 0.899 & \textbf{0.941} & 0.934 & 0.8927 \\
\hline
&  &  &  &  &  &\\[-5.5pt]
\tabincell{c}{U-Ones+CT\\+LSR \cite{pham2020interpreting}} & 0.825 & 0.855 & 0.937 & 0.930 & 0.923 & 0.8940 \\
\hline
\hline
&  &  &  &  &  &\\[-4.5pt]
Ours-RN50 & 0.897 & 0.838 & 0.893 & 0.932 & \textbf{0.938} & 0.8996 \\
\hline
&  &  &  &  &  &\\[-4.5pt]
Ours-RN152 & \textbf{0.920} & \textbf{0.886} & 0.907 & 0.937 & 0.933 & \textbf{0.9166} \\
\hline
\end{tabular}
}
\end{table}

\begin{table*}[!t]
\newcommand{\tabincell}[2]{\begin{tabular}{@{}#1@{}}#2\end{tabular}}
\renewcommand{\arraystretch}{0.65}
\caption{Results at various $T(IoU)$ from different models in the validation set of CheXpert. ``Ours" denotes the models trained without any box annotations. ``Ours$_{extra}$" denotes the models trained with extra box annotations from 457 images.}
\label{table:chexpert:tiou}
\centering
\scalebox{0.8}{
\begin{tabular}{c|c|c|c|c|c|c|c|c|c|c|c}
\hline
& & &  &  &  &  &  &  &  & & \\[-5.0pt]
 $T(IoU)$ & Model & Atelectasis & Cardiomegaly & Consolidation & Edema & \tabincell{c}{Enlarged \\Cardiomediastinum} & Pneumonia & Pneumothorax & \tabincell{c}{Pleural \\Effusion} & Fracture & Mean \\
\hline
\hline
& & &  &  &  &  &  &  &  & & \\[-5.0pt]
\multirow{5}{*}{0.1} & RN50 & 0.57 & 0.93 & 0.70 & 0.91 & 0.86 & 0.85 & 0.46 & 0.81 & 0.12 & 0.69 \\
\cline{2-12}
& & &  &  &  &  &  &  &  & & \\[-5.0pt]
& DRN50 & 0.58 & 0.66 & 0.69 & 0.73 & 0.50 & 0.88 & 0.26 & 0.78 & 0.19 & 0.59 \\
\cline{2-12}
& & &  &  &  &  &  &  &  & & \\[-5.0pt]
& Ours & 0.71 & 0.91 & 0.81 & 0.92 & 0.98 & 0.92 & 0.54 & 0.80 & 0.11 & 0.74 \\
\cline{2-12}
& & &  &  &  &  &  &  &  & & \\[-5.0pt]
& Ours$_{extra}$ & \textbf{0.79} & \textbf{0.99} & \textbf{0.85} & \textbf{0.99} & \textbf{1.00} & \textbf{0.95} & \textbf{0.74} & \textbf{0.95} & 0.22 & \textbf{0.83} \\
\cline{2-12}
& & &  &  &  &  &  &  &  & & \\[-5.0pt]
& Ours$_{extra}$ + SR & 0.71 & \textbf{0.99} & 0.82 & \textbf{0.99} & 0.99 & 0.93 & 0.71 & 0.94 & \textbf{0.26} & 0.82 \\
\hline
\hline
& & &  &  &  &  &  &  &  & & \\[-5.0pt]
\multirow{5}{*}{0.3} & RN50 & 0.16 & 0.49 & 0.31 & 0.53 & 0.34 & 0.43 & 0.25 & 0.43 & 0.02 & 0.33 \\
\cline{2-12}
& & &  &  &  &  &  &  &  & & \\[-5.0pt]
& DRN50 & 0.13 & 0.02 & 0.25 & 0.07 & 0.02 & 0.35 & 0.06 & 0.33 & 0.02 & 0.14 \\
\cline{2-12}
& & &  &  &  &  &  &  &  & & \\[-5.0pt]
& Ours & 0.21 & 0.35 & 0.43 & 0.52 & 0.37 & 0.62 & 0.16 & 0.35 & 0.02 & 0.34 \\
\cline{2-12}
& & &  &  &  &  &  &  &  & &  \\[-5.0pt]
& Ours$_{extra}$ & \textbf{0.36} & \textbf{0.99} & \textbf{0.48} & \textbf{0.85} & \textbf{0.98} & \textbf{0.63} & \textbf{0.40} & \textbf{0.70} & 0.04 & \textbf{0.60} \\
\cline{2-12}
& & &  &  &  &  &  &  &  & & \\[-5.0pt]
& Ours$_{extra}$ + SR & 0.26 & \textbf{0.99} & 0.38 & 0.71 & \textbf{0.98} & 0.56 & 0.32 & 0.51 & \textbf{0.05} & 0.53 \\
\hline
\hline
& & &  &  &  &  &  &  &  & & \\[-5.0pt]
\multirow{5}{*}{0.5} & RN50 & 0.02 & 0.10 & 0.08 & 0.08 & 0.02 & 0.11 & 0.09 & 0.11 & 0.00 & 0.07\\
\cline{2-12}
& & &  &  &  &  &  &  &  & &  \\[-5.0pt]
& DRN50 & 0.02 & 0.00 & 0.02 & 0.00 & 0.00 & 0.06 & 0.00 & 0.05 & 0.00 & 0.02 \\
\cline{2-12}
& & &  &  &  &  &  &  &  & & \\[-5.0pt]
& Ours & 0.04 & 0.03 & 0.11 & 0.07 & 0.05 & 0.13 & 0.05 & 0.06 & 0.00 & 0.06 \\
\cline{2-12}
& & &  &  &  &  &  &  &  & & \\[-5.0pt]
& Ours$_{extra}$ & \textbf{0.07} & \textbf{0.93} & \textbf{0.16} & \textbf{0.44} & \textbf{0.81} & \textbf{0.17} & \textbf{0.10} & \textbf{0.20} & 0.00 & \textbf{0.32} \\
\cline{2-12}
& & &  &  &  &  &  &  &  & & \\[-5.0pt]
& Ours$_{extra}$ + SR & 0.03 & 0.88 & 0.06 & 0.27 & 0.79 & 0.11 & 0.08 & 0.10 & 0.00 & 0.26 \\
\hline
\end{tabular}
}
\end{table*}

\begin{table*}[!t]
\newcommand{\tabincell}[2]{\begin{tabular}{@{}#1@{}}#2\end{tabular}}
\renewcommand{\arraystretch}{0.65}
\caption{Results at various $T(IoR)$ from different models in the validation set of CheXpert.}
\label{table:chexpert:tior}
\centering
\scalebox{0.8}{
\begin{tabular}{c|c|c|c|c|c|c|c|c|c|c|c}
\hline
& & &  &  &  &  &  &  &  & & \\[-5.0pt]
 $T(IoR)$ & Model & Atelectasis & Cardiomegaly & Consolidation & Edema & \tabincell{c}{Enlarged \\Cardiomediastinum} & Pneumonia & Pneumothorax & \tabincell{c}{Pleural \\Effusion} & Fracture & Mean \\
\hline
\hline
& & &  &  &  &  &  &  &  & & \\[-5.0pt]
\multirow{5}{*}{0.1} & RN50 & 0.59 & 0.98 & 0.75 & 0.95 & 0.98 & 0.90 & 0.55 & 0.83 & 0.15 & 0.74 \\
\cline{2-12}
& & &  &  &  &  &  &  &  & & \\[-5.0pt]
& DRN50 & 0.62 & 0.95 & 0.78 & 0.90 & 0.92 & 0.94 & 0.56 & 0.84 & 0.22 & 0.75 \\
\cline{2-12}
& & &  &  &  &  &  &  &  & & \\[-5.0pt]
& Ours & 0.73 & 0.98 & 0.87 & 0.97 & 0.99 & 0.96 & 0.64 & 0.84 & 0.12 & 0.79 \\
\cline{2-12}
& & &  &  &  &  &  &  &  & & \\[-5.0pt]
& Ours$_{extra}$ & \textbf{0.83} & \textbf{0.99} & 0.88 & \textbf{1.00} & \textbf{1.00} & \textbf{0.97} & \textbf{0.85} & \textbf{0.97} & 0.26 & 0.86 \\
\cline{2-12}
& & &  &  &  &  &  &  &  & & \\[-5.0pt]
& Ours$_{extra}$ + SR & 0.82 & \textbf{0.99} & \textbf{0.90} & \textbf{1.00} & \textbf{1.00} & \textbf{0.97} & \textbf{0.85} & \textbf{0.97} & \textbf{0.32} & \textbf{0.87} \\
\hline
\hline
& & &  &  &  &  &  &  &  & & \\[-5.0pt]
\multirow{5}{*}{0.5} & RN50 & 0.06 & 0.93 & 0.27 & 0.73 & 0.90 & 0.47 & 0.36 & 0.29 & 0.00 & 0.45 \\
\cline{2-12}
& & &  &  &  &  &  &  &  & & \\[-5.0pt]
& DRN50 & 0.07 & 0.89 & 0.48 & 0.67 & 0.73 & 0.58 & 0.39 & 0.40 & 0.02 & 0.47 \\
\cline{2-12}
& & &  &  &  &  &  &  &  & & \\[-5.0pt]
& Ours & 0.12 & 0.94 & 0.49 & 0.71 & 0.91 & 0.65 & 0.37 & 0.35 & 0.02 & 0.51 \\
\cline{2-12}
& & &  &  &  &  &  &  &  & &  \\[-5.0pt]
& Ours$_{extra}$ & 0.40 & \textbf{0.99} & 0.56 & 0.88 & \textbf{1.00} & 0.74 & 0.54 & 0.73 & 0.03 & 0.66 \\
\cline{2-12}
& & &  &  &  &  &  &  &  & & \\[-5.0pt]
& Ours$_{extra}$ + SR & \textbf{0.50} & \textbf{0.99} & \textbf{0.65} & \textbf{0.92} & \textbf{1.00} & \textbf{0.85} & \textbf{0.63} & \textbf{0.85} & \textbf{0.08} & \textbf{0.72} \\
\hline
\hline
& & &  &  &  &  &  &  &  & & \\[-5.0pt]
\multirow{5}{*}{0.75} & RN50 & 0.01 & 0.83 & 0.06 & 0.48 & 0.79 & 0.26 & 0.20 & 0.08 & 0.00 & 0.30 \\
\cline{2-12}
& & &  &  &  &  &  &  &  & &  \\[-5.0pt]
& DRN50 & 0.01 & 0.83 & 0.27 & 0.48 & 0.60 & 0.36 & 0.31 & 0.17 & 0.01 & 0.34 \\
\cline{2-12}
& & &  &  &  &  &  &  &  & & \\[-5.0pt]
& Ours & 0.02 & 0.82 & 0.30 & 0.46 & 0.70 & 0.48 & 0.23 & 0.15 & 0.00 & 0.35 \\
\cline{2-12}
& & &  &  &  &  &  &  &  & & \\[-5.0pt]
& Ours$_{extra}$ & 0.17 & 0.89 & 0.35 & 0.65 & 0.95 & 0.59 & 0.35 & 0.43 & 0.01 & 0.54 \\
\cline{2-12}
& & &  &  &  &  &  &  &  & & \\[-5.0pt]
& Ours$_{extra}$ + SR & \textbf{0.30} & \textbf{0.94} & \textbf{0.46} & \textbf{0.78} & \textbf{0.98} & \textbf{0.61} & \textbf{0.48} & \textbf{0.61} & \textbf{0.04} & \textbf{0.58} \\
\hline
\end{tabular}
}
\end{table*}

\subsection{CheXpert Dataset}
CheXpert is another prominent CXR dataset containing 224,316 chest radiographs of 65,240 patients. However, the images in CheXpert are annotated with labels only at image level.
To further illustrate the localization performance on this dataset, we invite a senior radiologist with 10+ years of experience to label the bounding boxes for 9 abnormalities, \eg, ``Atelectasis", ``Cardiomegaly", ``Consolidation", ``Edema", ``Enlarged Cardiomediastinum", ``Pneumonia", ``Pneumothorax", ``Pleural Effusion", and ``Fracture". In the end, 2345 images were annotated with 6099 bounding boxes for the 9 abnormalities. It is worth noting that the number of our box annotations is significantly larger than the number of annotated boxes in the NIH dataset.
These new box annotations on the CheXpert dataset will be released soon.

\subsubsection{Abnormality classification}
The authors of CheXpert \cite{irvin2019chexpert} propose an evaluation protocol over 5 categories: ``Atelectasis", ``Cardiomegaly", ``Consolidation", ``Edema", and ``Pleural Effusion", which were selected based on the clinical importance and prevalence from the validation set. In this experiment, we use the official set to train all the models, and show the AUC scores of these 5 abnormalities on official validation set. 
 CheXpert captures uncertainties inherent in radiograph interpretation with an effective labeling strategy ($0$ for negative, $-1$ for uncertain, and $1$ for positive). There are a few significant differences between the performance of the uncertainty approaches.
\textit{U-Ignore} \cite{irvin2019chexpert} ignores the uncertainty labels during training, while \textit{U-Zeros} \cite{irvin2019chexpert}  and \textit{U-Ones} \cite{irvin2019chexpert} treat them as $0$ or $1$.
 Since the \textit{U-Ones} model achieves the best AUC performance in this dataset, we treat all uncertainty labels as $1$ in our experiments. At the same, we show the results of our methods with two backbone networks (ResNet50 and ResNet152). 
 AUC scores are shown in Table \ref{table:auc:chexpert}.
Our method outperforms the three models in the official paper \cite{irvin2019chexpert}, indicating that our method can maintain competitive classification performance.

In Table \ref{table:auc:chexpert}, we also compare the performance of DenseNet-121 with the conditional training and label smoothing regularization (\textit{U-Ones+CT+LSR}) strategies in \cite{pham2020interpreting}. Pham \etal \cite{pham2020interpreting} obtained mean AUC score of 0.930 on the official CheXpert testing set with the ensemble approach that combined six deep models of DenseNet-121, DenseNet-169, DenseNet-201, Inception-ResNet-v2, Xception, and NASNetLarge. In \cite{pham2020interpreting}, the DenseNet-121 with the strategies of \textit{U-Ones+CT+LSR} is the single model that achieves the best performance in the official CheXpert validation set. In comparison, our method can outperform DenseNet-121 with the strategies of \textit{U-Ones+CT+LSR} even with smaller backbone of ResNet50 on the same validation set. The mean AUC scores on the official test set of our method with the backbone of ResNet50 and ResNet152 are 0.888 and 0.895 respectively without the implementation of an ensemble strategy. 
 

\begin{figure}[!t]
\centering
\includegraphics[width=8.5cm, height=10.8cm]{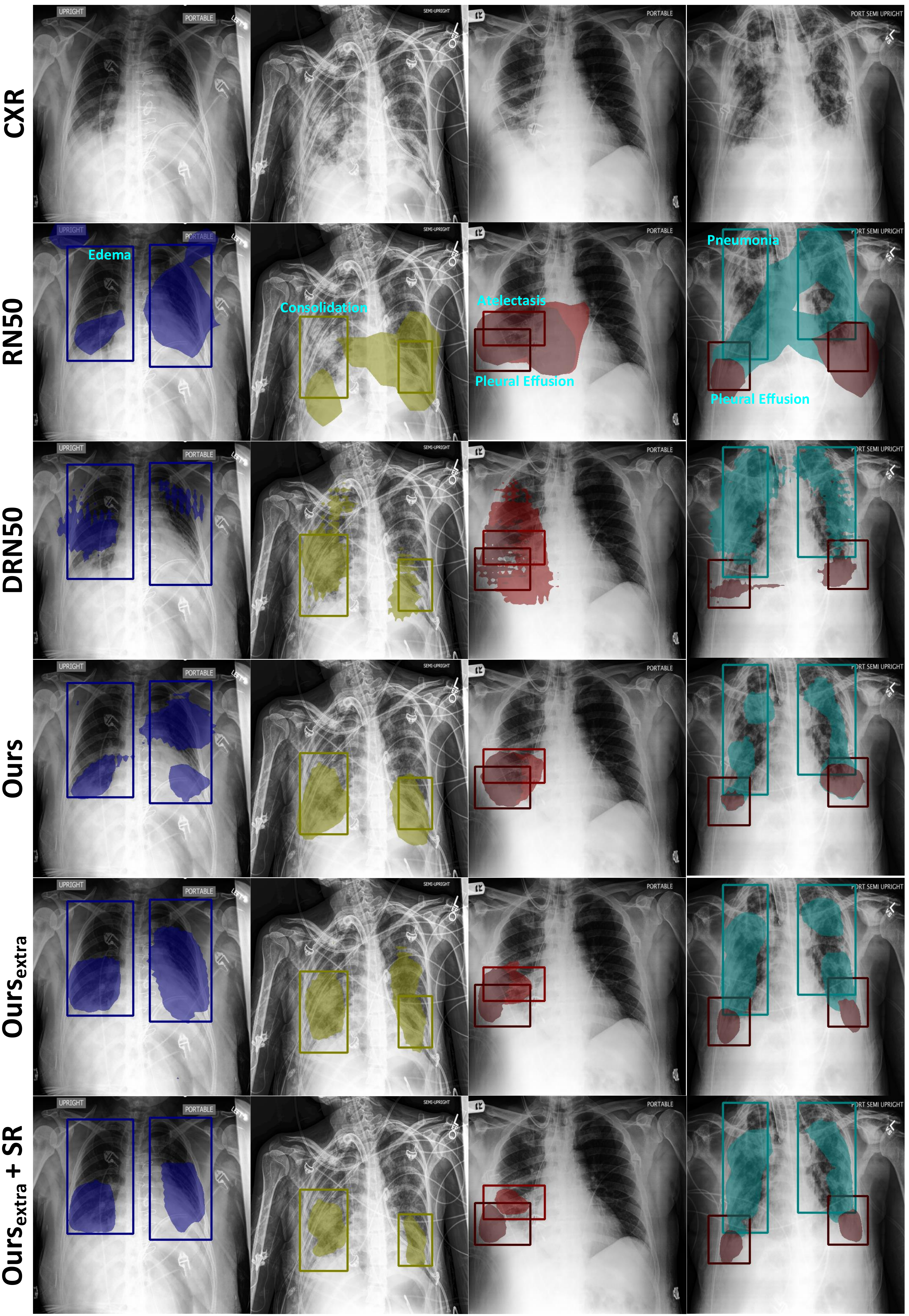}
\caption{Qualitative comparison of results between different models. We compare the attention results of four cases from the two baseline models and our methods.}
\label{fig:chexpert-vis}
\end{figure}

\subsubsection{Abnormality localization}
In this experiment, we split 221,674 images from the official training set to train the models, which including 457 images with 1435 bounding boxes. Then, we split 1888 images with 4664 bounding boxes as the validation set to evaluate the localization performance of 9 abnormalities. Table \ref{table:chexpert:tiou} and Table \ref{table:chexpert:tior} show the localization results in the validation set. Here, we compare with two baseline models, ResNet50 and Dilated ResNet50. Dilated ResNet50 encodes the size of feature map to $\frac{1}{8}$ of the input images and uses LSE pooling. We use CAM \cite{zhou2016learning} to generate the attention maps as the localization results for two baseline methods. We can see from these tables that our method outperforms both these baseline methods. Moreover, with the addition of very few extra annotations (457 images), our method (``Ours$_{extra}$") can gain a significant improvements with respect to localization scores. We also show the results of self-refinement method (``Ours$_{extra}$ + SR"), which can improve the $T(IoR)$ results.

Qualitative results are shown in Fig. \ref{fig:chexpert-vis}, where we observe that our method can generate more accurate localization results. We can see the baseline ResNet50 model usually produces rough and large attention maps, while our method can produce more accurate results. The gridding effect can be observed in Fig. \ref{fig:chexpert-vis} from the results of the Dilated ResNet50 model, which is caused by dilated convolution \cite{wang2018understanding}. We can see that our method can eliminate this effect when using dilated convolution to increase the resolution of attention maps. With the addition of very few extra annotations, we can see that attention results of our method can be further improved. We can also see that the attention results using the self-refinement method are concentrated within the box annotations and look anatomically plausible and appear close to the image segmentation effect.

\section{Discussion and Conclusion}
We presented a novel hierarchical attention framework comprised of activation- and gradient-based attention mechanisms to address the CXR image diagnosis and corresponding abnormality localization problems. We evaluated our method on two public datasets and compared with recent state-of-the-art methods. Extensive experimental results show that our method can achieve state-of-the-art results on both image-level classification and abnormality localization tasks. Our method can be easily generalized to other weakly-supervised problems with limited box- or pixel-level annotations. Furthermore, our localization results can be used by radiologists for verifying diagnosis conclusions, thereby providing direct relevance in a clinical environment. These visual cues can also be used in an active learning framework where the radiologist can guide the model towards improved predictions, thereby helping infuse human domain knowledge in building continually-learning algorithms.

We next briefly discuss the limitations of our proposed method. First, the extension to 3D medical images may not be directly feasible. Because 3D medical images contain richer details of anatomical and pathological cues, 3D generalization from the 2D design of our attention modules may not be trivial. The implementation of 3D self-attention modules and 3D CAM/Grad-CAM may also need to consider the efficiency of GPU usage. Meanwhile, since some of 3D medical images, \eg, Computer Tomography (CT), may have the lower resolution in the z-direction, the design of 3D self-attention and 3D Grad-CAM/CAM may need to consider the factor of anisotropic resolution. Further, the definition of label hierarchy for 3D images may also be very distinctive to the label hierarchy of 2D images.

Second, most abnormalities involved in this work are related to soft tissues. As can be found in Tables \ref{table:chexpert:tiou} and \ref{table:chexpert:tior}, the performances for bone tissue abnormality of fracture from all methods are not very promising. Bone-tissue abnormalities like fractures are relatively subtle and can appear in a thin elongated shape. Accordingly, the detection of these kinds of abnormalities may need to incorporate the constraint of anatomical structures for reducing the search space. Since only two levels of label hierarchy is exploited in this study without the explicit consideration of broader anatomical labels of rib, scapula, spine, \etc, our current model may not be sufficient to address the localization of the difficult abnormality of fracture.

Third, there also exist fine-grained hierarchies of abnormality labels in the NIH Chest-Xray14 and CheXpert, which may also be informative for improving performance. For example, pneumonia is the most common cause of lung consolidation, and has some related complications such as abscesses, pleural effusion and infiltration. Such fine-grained hierarchies could be helpful for learning sharable features or to some degree mitigate the label/sample imbalance problem. As part of future work, we will further explore the broader and deeper hierarchies of abnormality and anatomy labels with the specific design of ordinal constraints in the hierarchical attention framework.

Finally, we notice there exist some questions for the box annotations of some abnormalities (\eg, ``Effusion", ``Infiltration" and ``Pneumonia") in NIH Chest Xray14. We have invited an experienced radiologist to carefully review the original annotated boxes for these abnormalities and give a detailed analysis in the supplementary material. It can be observed that some boxes tend to cover more non-related regions, or there are still some missing boxes. In such cases, even though our results deliver higher quality in terms of clinical findings, they still cannot better match with the boxes of NIH Chest Xray14. It leads to an inaccurate comparison with different methods in high $T(IoU)$ thresholds. To better solve this issue, we hope to invite several senior radiologists to perform the task of mask-level annotation for NIH and CheXpert dataset in the future. These precise mask-level annotations can be used to conduct a more meaningful localization comparison at high $T(IoU)$ thresholds.

\normalem

\clearpage
\renewcommand {\thefigure} {S\arabic{figure}}
\renewcommand {\thetable} {S\arabic{table}}
\section{Supplementary Material}
\subsection{Analysis of Box Annotations in NIH Chest X-ray14}
Referring to Tables III, IV, and V in the paper, our results are generally better than the compared methods of Li \etal [12] and Liu \etal [21]. In particular, at the settings of $T(IoU) = 0.1$ in Table III, $T(IoU) < 0.5$ in Table IV, and $T(IoR) <= 0.5$ in Table V, our method can remarkably improve the localization performance for most abnormality classes, especially for ``Nodule”. 
Accordingly, the efficacy of our method can be corroborated. 
However, for some diffusive and fuzzy abnormalities (\eg, ``Effusion", ``Infiltration", ``Pneumonia"), our method does not produce great scores in high $T(IoU)$ thresholds. We have invited an experienced radiologist to carefully review the originally annotated boxes in NIH Chest X-ray14 and suggest that the variety of the annotation coverage for these classes is very large (see Fig. \ref{fig:problems} for illustration).

\begin{figure}[h]
\setcounter{figure}{0}
\centering
\includegraphics[width=8.4cm, height=6.9cm]{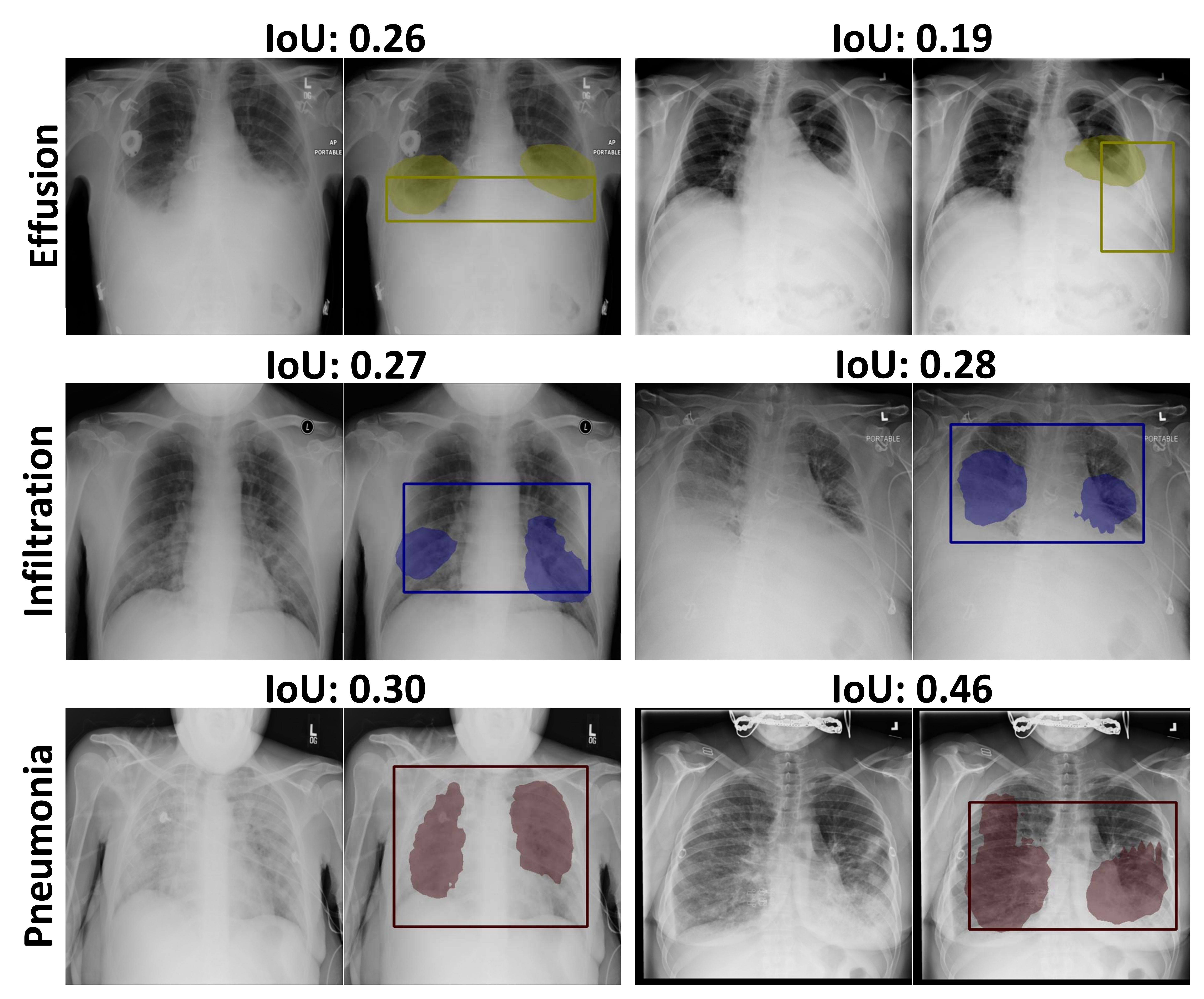}
\caption{Cases of ``Effusion", ``Infiltration", and ``Pneumonia" with rough ground-truth boxes in NIH Chest-Xray14. We show the box annotations of the NIH Chest-Xray14, our localization masks, and the corresponding IoU scores. For these cases with ``Effusion", ``Infiltration" and ``Pneumonia" in two lungs, the box annotations are drawn across two lungs even with the mediastinum included, which may be problematic for evaluation. Our localization results are closer to the abnormal regions, but cannot better match with the boxes enclose many non-related regions.}
\label{fig:problems}
\end{figure}

\begin{figure}[!t]
\centering
\includegraphics[width=8.5cm, height=2.3cm]{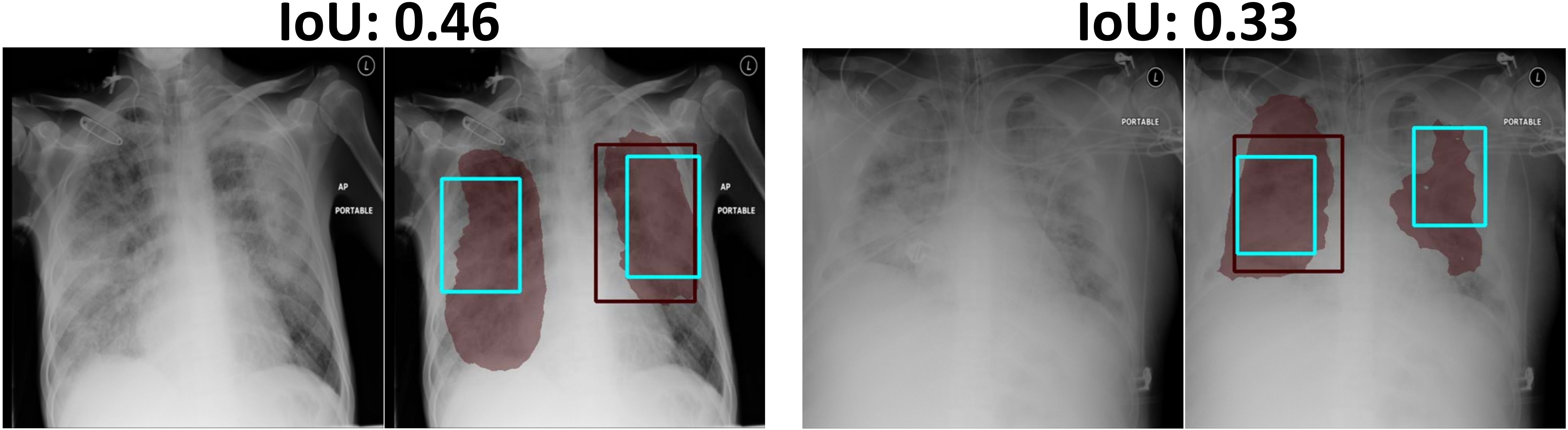}
\caption{Problems of annotations of two ``Pneumonia" cases in NIH Chest-Xray14. We present the boxes of NIH Chest-Xray14 in brown boxes, and the boxes from our invited senior radiologist in cyan. The localization results from our method are presented in brown masks. We also show the IoU scores of our localization results and the boxes of NIH Chest-Xray14.}
\label{fig:pneumonia-problems}
\end{figure}

As can be found in Fig. \ref{fig:problems}, the annotated boxes may sometimes enclose many non-related regions. For such cases, if the abnormalities are presented in both lungs, the boxes may be drawn roughly to even include the mediastinum. The inclusion of mediastinum may be problematic as lung infiltration or pneumonia may not happen there. The relatively large and rough boxes are not favor our results if the setting of $T(IoU)$ is higher. Since our results may be more exact to localize the abnormalities, our results for these abnormalities have less chance to be present at mediastinum, and therefore, may not perfectly match with the rough box in high IoU scores. It can be observed that the IoU scores of all the cases in Fig. \ref{fig:problems} are around from 0.1 to 0.5. This may explain the lower scores at high $T(IoU)$ thresholds in Table III and IV. 

Additionally, for ``Pneumonia", our invited senior radiologist also provide some relabeled bounding boxes and we show them in Fig. \ref{fig:pneumonia-problems}. Although most boxes provided by NIH Chest Xray14 for ``Pneumonia" are reasonable, it is worth noting that there are some missing boxes. As can be found in the figure, our method can also identify the ``Pneumonia" regions where the boxes were missed by NIH annotators. In such cases, our results may not hold high IoU scores, but indeed deliver higher quality in terms of clinical findings. Due to the above, we suggest the results with IoU scores in the range of 0.3 to 0.7 are acceptable (see ``Pneumonia" in Fig. \ref{fig:problems} and Fig. \ref{fig:pneumonia-problems}). The performances of our method for the ``Pneumonia" class are reported at the settings of $T(IoU)=[0.1, 0.2, 0.3, 0.4, 0.5, 0.6, 0.7]$ in Table IV. As can be found, more than 65\% cases of our results have IoU scores larger than 0.3, whereas there are approximate 34\% ($0.69-0.35$) and 29\% ($0.35-0.06$) cases of our results with IoU scores between 0.3 to 0.5, and 0.5 to 0.7, respectively. And 6\% of our results have more than 0.7 IoU scores with the boxes. With such distribution, the major portions of our results, nearly 63\% ($34\%+29\%$) has IoU scores in the range of 0.3 to 0.7, compared to 24\% ($0.49-0.25$) of results in Liu \etal [21] with the IoU score range of 0.3 to 0.7. 

\end{document}